\documentclass{article}
\pdfoutput=1
\usepackage[final, nonatbib]{neurips_2021}

\usepackage[utf8]{inputenc} % allow utf-8 input
\usepackage[T1]{fontenc}    % use 8-bit T1 fonts
\usepackage{hyperref}       % hyperlinks
\usepackage{url}            % simple URL typesetting
\usepackage{booktabs}       % professional-quality tables
\usepackage{amsfonts}       % blackboard math symbols
\usepackage{nicefrac}       % compact symbols for 1/2, etc.
\usepackage{microtype}      % microtypography
\usepackage{xcolor}         % colors
\usepackage{color}
\usepackage{amsmath,amsfonts,amsthm} % Math packages
\usepackage{algorithm}
\usepackage{algorithmicx}
\usepackage{algpseudocode} 
\usepackage{colortbl}
\usepackage{wrapfig}
\usepackage{graphicx}
\usepackage{subfigure}
\usepackage{amssymb}
\usepackage{mathrsfs}
\pdfoutput=1

\definecolor{gychen}{rgb}{0,0,0} 
\definecolor{drd}{rgb}{0,0,0}

\title{Flattening Sharpness for Dynamic Gradient Projection Memory Benefits Continual Learning}

\author{
  Danruo Deng${^1}$,~
  Guangyong Chen$^{2}$\thanks{Corresponding author: Guangyong Chen. <gy.chen@siat.ac.cn>},~
  Jianye Hao$^{3,4}$,~
  Qiong Wang$^{2}$,~
  Pheng-Ann Heng$^{1,2}$ \\
  ${^1}$The Chinese University of Hong Kong,
   ${^2}$Guangdong-Hong Kong-Macao Joint Laboratory \\ of Human-Machine Intelligence-Synergy Systems, Shenzhen Institute of Advanced Technology, \\ 
   Chinese Academy of Sciences, 
   ${^3}$College of Intelligence and Computing, Tianjin University, \\${^4}$Huawei Noah’s Ark Lab\\
   \texttt{\{drdeng,pheng\}@cse.cuhk.edu.hk, \{gy.chen, wangqiong\}@siat.ac.cn,} \\
   \texttt{jianye.hao@tju.edu.cn} 
}

\begin{document}

\maketitle

\begin{abstract}
\label{abs} 
The backpropagation networks are notably susceptible to catastrophic forgetting, where networks tend to forget previously learned skills upon learning new ones. To address such the 'sensitivity-stability' dilemma, most previous efforts have been contributed to minimizing the empirical risk with different parameter regularization terms and episodic memory, but rarely exploring the usages of the \textit{weight loss landscape}. In this paper, we investigate the relationship between the weight loss landscape and sensitivity-stability in the continual learning scenario, based on which, we propose a novel method, \textit{Flattening Sharpness for Dynamic Gradient Projection Memory (FS-DGPM)}. In particular, we introduce a soft weight to represent the importance of each basis representing past tasks in GPM, which can be adaptively learned during the learning process, so that less important bases can be dynamically released to improve the sensitivity of new skill learning. We further introduce \textit{Flattening Sharpness (FS)} to reduce the generalization gap by explicitly regulating the flatness of the weight loss landscape of all seen tasks. As demonstrated empirically, our proposed method consistently outperforms baselines with the superior ability to learn new skills while alleviating forgetting effectively.\textcolor{drd}{\footnote{The code is available at: \url{https://github.com/danruod/FS-DGPM}}.}

%The backpropagation networks are notably susceptible to catastrophic forgetting, where networks tend to completely forget previously learned skills upon learning new ones. To address such 'sensitivity-stability' dilemma, most previous efforts have been contributed to adopting parameter regularization and episodic memory, but only achieving limited practical performance. This paper is motivated by that the interference between sequentially learned tasks is minimized if the input vectors are orthogonal to each other. However, the different dimensions of the input vector can not be treated equally due to the heterogeneity of data. To address this problem, we introduce a soft weight to represent the importance of each dimension, which can be adaptively learned by the meta-learning algorithm during the learning process, so that less important dimension of previous tasks can be dynamically released to ensure the ability to learn new skills. We further introduce a regularization term in the meta-objective, so that the cross-task gradients are constrained to be orthogonal to the important gradient space. As demonstrated empirically, our proposed method consistently outperforms the state-of-the-art method with the superior ability to learn new skills while reducing forgetting effectively.

\end{abstract}
  
\section{Introduction}
\label{intro}

Humans have the ability to continually learn new knowledge without forgetting their previously learned ones through mediating a rich set of neurocognitive mechanisms \cite{wilson1994reactivation,ji2007coordinated,van2020brain}. This ability, often known as continual learning or lifelong learning \cite{parisi2019continual}, is crucial for computational systems, such as deep neural networks (DNNs), which are required to sequentially learn and deal with multiple tasks when implemented in the dynamically changing environment. Continual learning remains a long-standing challenge for DNNs since these networks are typically trained with stationary training batches by stochastic gradient descent methods \cite{le2011optimization}, which generally leads to an abrupt performance decrease on previously learned tasks as new tasks are learned. To address such \textit{catastrophic forgetting}, we can brutally retrain an oracle network on the entire dataset containing all tasks to capture dynamic changes in the data distribution, but this methodology is obviously too inefficient to hinder the learning of novel data in real time.

During the last few years, lots of research efforts have been devoted to improving the \textit{stability} of DNNs on old tasks while keeping \textit{sensitive} to new information. The first intuitive idea is to introduce an independent branch for each new task while freezing the old task parameters to preserve the old knowledge \cite{rusu2016progressive,yoon2017lifelong,xu2018reinforced,mallya2018packnet,serra2018overcoming,li2019learn}. However, in this way, the network will inevitably become redundant as the task number continually increases. As presented in the neurocognitive works \cite{wilson1994reactivation,ji2007coordinated}, the reactivation of neuronal activity patterns, representing old memories, plays an important role in the continual learning of humans \cite{van2020brain}. Thus, forgetness can be effectively mitigated by training a single network for new tasks by considering diverse information stored in the memory, including the original training samples of old tasks \cite{riemer2018learning,chaudhry2019continual,gupta2020maml}, the gradients induced from old tasks \cite{farajtabar2020orthogonal} and the feature subspace representing old tasks \cite{saha2021gradient}. However, their continual learning performance is still limited because DNNs can easily overfit the limited information stored in the small-size memory.

The overfitting problem of DNNs is often attributed to the complex loss landscape containing multiple local optima, and the sharpness of the loss landscape has been widely used to characterize the generalization gap in standard training scenarios from both theoretical and empirical perspectives \cite{neyshabur2017exploring,li2018visualizing,wu2020adversarial,foret2020sharpness,chen2020noise}. While this characterization has inspired new approaches for model training with better generalization, practical algorithms that especially seek out flatter minima to effectively address the 'sensitivity-stability' dilemma for continual learning have thus far been elusive. In this paper, our first contribution is to characterize the weight loss landscape for the continual learning scenario and identify that a flatter loss landscape with lower loss value often leads to better continual learning performance, as shown in Figure \ref{fig_landscape} and Figure \ref{fig_fser}. 

Further, based on our characterization of the weight loss landscape, we find that the recently proposed Gradient Projection Memory (GPM) method \cite{saha2021gradient} maintains the lowest loss value on old tasks among the previously proposed methods by taking gradient steps orthogonal to the subspace representing old tasks. However, its loss landscape on newly learned tasks is the sharpest due to the lack of sufficient subspace left for new task learning. To improve the network's sensitivity, our second contribution is to predict the importance of bases spanning the subspace for old tasks, so that less important bases can be dynamically released. In particular, we introduce a soft weight to indicate the bases importance, which can be dynamically adjusted by combining the \textit{Flattening Sharpness (FS)} to minimize the loss value and loss sharpness simultaneously.
%which can be dynamically solved by minimizing the loss value and loss sharpness simultaneously. 
Intuitively, a basis will be regarded as important for preserving old knowledge if the gradients induced by new tasks and old ones are aligned in the opposite direction on that basis. As demonstrated through extensive experiments, our proposed method can consistently outperform the state-of-the-art methods \cite{kirkpatrick2017overcoming,rebuffi2017icarl, lopez2017gradient, riemer2018learning, chaudhry2019continual, gupta2020maml,saha2021gradient} by a notable margin across a range of widely used benchmark datasets.

\section{Related Work}
\label{sec_related}

In this section, we briefly survey the representative works of continual learning by highlighting their contributions. To simplify our presentation, this section is organized by dividing the representative works into three categories, parameter isolation, regularization-based, memory-based methods. 

\textbf{Parameter isolation} methods address forgetting by assigning a different subset of network parameters to each task. Without restrictions on network architecture, new neurons or layers or modules can be added for new tasks, while the previous task parameters can be frozen or copied to preserve old knowledge. For instance, Progressive Neural Network (PGN) \cite{rusu2016progressive} freezes the parameters trained with previous knowledge while expands the architecture by allocating new sub-networks with fixed capacity for new tasks. Dynamically Expandable Networks (DEN) \cite{yoon2017lifelong} selectively retrains or expands network capacity by splitting/duplicating important units on new tasks. Reinforced Continual Learning (RCL) \cite{xu2018reinforced} uses reinforcement learning strategy to adaptively expand the network of each layer, while \cite{li2019learn} use neural architecture search to find optimal network structures for each sequential task. Alternatively, with the architecture remaining static, a fixed part is allocated to each task. During the training of a new task, previous task parts are masked out to prevent interference. The mask sets are imposed at parameter level \cite{mallya2018packnet}, or unit level \cite{serra2018overcoming}. PackNet \cite{mallya2018packnet}  uses iterative pruning to fully restrict gradient updates on important weights via a binary mask, whereas HAT \cite{serra2018overcoming} limits the update of important units recognized by the hard attention mask through gradient backpropagation.

\textbf{Regularization-based} methods introduce an additional regularization term in the loss function to consolidate previous knowledge without using replay. This involves using knowledge distillation \cite{li2017learning, hu2018overcoming} or penalizing changes in weights deemed important for previous tasks \cite{kirkpatrick2017overcoming, zenke2017continual, nguyen2017variational, aljundi2018memory, cha2020cpr} to reduce forgetting. There are many ways to measure the importance. Elastic Weight Consolidation (EWC) \cite{kirkpatrick2017overcoming} identifies important weights based on the diagonal values of Fisher information matrix after training, while Synaptic Intelligence (SI) \cite{zenke2017continual} calculates them online and over the entire learning trajectory in parameter space. Memory Aware Synapses (MAS) \cite{aljundi2018memory} estimates importance weights in an unsupervised manner, while Variational Continual Learning (VCL) \cite{nguyen2017variational} introduces a variational framework that spawned some Bayesian-based works \cite{ritter2018online, ebrahimi2019uncertainty, ahn2019uncertainty, chen2019facilitating}. For example, \cite{ritter2018online} recursively uses a Gaussian Laplace approximation of the Hessian to approximate the posterior after every task, \cite{ebrahimi2019uncertainty} adjusts the learning rate according to the uncertainty defined by the probability distribution of the network weights. \cite{ahn2019uncertainty} introduces an interpretation of node-wise uncertainty on the Kullback-Leibler (KL) divergence term of the variational lower bound for Gaussian mean-field approximation. 

%\textcolor{drd}{Recently proposed stable SGD \cite{mirzadeh2020understanding} and CPR \cite{cha2020cpr} have a similar point of view with ours on the advantages of wide local minima. However, stable SGD mainly focuses on the effect of training regimes and CPR adds classifier projection as a regularization terms to promote wide local minima respectively.}
% \textcolor{drd}{Recently proposed stable SGD \cite{mirzadeh2020understanding} and CPR \cite{cha2020cpr} have similar views with ours on the advantages of wide local minima. However, stable SGD mainly focuses on the effect of training regimes, such as learning rate, batch size and drop out. CPR adds classifier projection as a regularization terms to promote wide local minima. Moreover, stable SGD and CPR analysis of loss landscape is limited to one task.}

Our method mainly follows \textbf{memory-based} methods, which mitigate forgetting based on information extracted from old tasks or based on a generative model to generate pseudo samples. For example, iCaRL \cite{rebuffi2017icarl} selects and stores samples closest to the feature mean of each class. ER \cite{chaudhry2019continual, riemer2018learning} suggests reservoir sampling under the limited and fixed budget for replay buffer. Deep Generative Replay (DGR) \cite{shin2017continual} trains a deep generative model in the Generative Adversarial Network (GAN) framework \cite{goodfellow2014generative} to simulate past data. These previous task samples are mainly reused as model inputs for replay in the above methods. However, replay might be prone to overfitting the subset of stored samples. Alternatively, samples stored in memory can also be used to constrain the optimization of the new task loss to prevent previous task interference, thereby leaving more leeway for backward and forward transfer. Gradient Episodic Memory (GEM) \cite{lopez2017gradient} projects the estimated gradients in the feasible region, which is outlined by previous task gradients calculated from the episodic memory samples. Averaged-GEM (A-GEM) \cite{chaudhry2018efficient} relaxes the projection to a direction that is estimated from samples randomly selected from memory. \cite{guo2020improved} proposes a unified view of episodic memory-based continual learning methods, including GEM and A-GEM, and improves performance over these methods by using a loss-balancing update scheme. A few other works have utilized gradient information to protect previous knowledge. \textcolor{gychen}{\cite{riemer2018learning, gupta2020maml} adopt optimization-based meta-learning to enforce gradient alignment between samples from different tasks}. GPM \cite{saha2021gradient} minimizes interference between sequential tasks by ensuring that gradient updates only occur in directions orthogonal to the input of previous tasks.
\section{The Weight Loss Landscape of Continual Learning}
\label{sec_pre}

%\textcolor{gychen}{I think it is better to introduce our motivation here, which is the connection between the landscape and the continual learning performance. We can follow Yisen's Neurips work, then our method can be introduced by solid experimental visualization results. \\ We should also give a simple case study by flattening sharpness on ER method. }

In this section, we first introduce our formulation of continual learning, and then characterize the weight loss landscape for the continual learning scenario from stability and sensitivity. Finally, some insights combining the weight loss landscape and continual learning are provided.

\subsection{Problem Formulation} 
Throughout the paper, we denote scalars as $a$, vectors as $\boldsymbol{a}$, matrices as $\boldsymbol{A}$, and sets as $\mathcal{A}$. We consider a supervised learning setup where $T$ tasks are sequentially learned from their training data. Each task has an identical task descriptor, $\tau\in\{1,2,\ldots,T\}$, with its dataset $\mathcal{D}_{\tau}=\{\boldsymbol{x}_{i,\tau},y_{i,\tau}\}_{i=1}^{n_{\tau}}$ containing $n_{\tau}$ samples randomly generated from a latent distribution $\mathscr{D}_{\tau}$. At any time-step during the learning process, we minimize the empirical risk of the model on all $t$ tasks seen so far, with just limited size of memory $\mathcal{M}$ to summarize the training data of previous tasks $\{\mathcal{D}_{\tau}\}_{\tau=1}^{t-1}$. To simplify the notation, we denote $L_{\mathcal{A}}(\boldsymbol{w}) = \frac{1}{|\mathcal{A}|}\sum_{\left(\boldsymbol{x}, y\right)\in \mathcal{A}}\left[\ell \left(f_{\boldsymbol{w}}\left(\boldsymbol{x} \right), y\right)\right]$ as the average empirical loss for the set $\mathcal{A}$, where $\ell(\cdot, \cdot)$ is an arbitrary loss function (\textit{e.g.} the cross-entropy (CE) loss), $|\mathcal{A}|$ is the sample 
size of the set $\mathcal{A}$, and $f_{\boldsymbol{w}}$ is the DNN with weight $\boldsymbol{w}$. Our final goal is to find an optimal parameter $\boldsymbol{w}$, which minimizes the overall risk $\sum_{\tau=1}^TL_{\mathscr{D}_{\tau}}(\boldsymbol{w})$ for all tasks.

\subsection{Connection of Weight Loss Landscape and Continual Learning}
\label{subsec_visual}
% After learning a new task, we visualize the weight loss landscape of each task seen so far by plotting changes in its training loss when moving the weights $\boldsymbol{w}$ in a random direction $\boldsymbol{d}$ with magnitude $\alpha$ following \cite{li2018visualizing}:
After learning a new task, we visualize the weight loss landscape of each task seen so far by plotting changes in its training loss when moving the weights $\boldsymbol{w}$ in a random direction $\boldsymbol{d}$ with magnitude $\alpha$ following \cite{li2018visualizing}:
\begin{equation}
\label{eq_visual}
g_t(\alpha) = L_{\mathcal{D}_t}(\boldsymbol{w}+\alpha \boldsymbol{d}) = \frac{1}{|\mathcal{D}_t|}\sum_{\left(\boldsymbol{x}, y\right)\in \mathcal{D}_t} \ell \left(f_{\boldsymbol{w}+\alpha\boldsymbol{d}}\left(\boldsymbol{x} \right), y\right),
\nonumber
\end{equation}
where $\mathcal{D}_t$ is the training set for the $t$-th task previously learned. To eliminate the scaling invariance of DNNs, $\boldsymbol{d}$ is sampled  from a Gaussian distribution and further normalized by $\boldsymbol{d}_{l,j} \leftarrow \frac{\boldsymbol{d}_{l,j}}{\|\boldsymbol{d}_{l,j}\|_F}\|\boldsymbol{w}_{l,j}\|_F$, where $\boldsymbol{d}_{l,j}$ represents the $j$-th filter at $l$-th layer of $\boldsymbol{d}$, and $\|\cdot\|_F$ denotes the Frobenius norm. \textcolor{drd}{Compared with our visualization,} \cite{mirzadeh2020understanding} only consider one task and plot the loss landscape along the directions of \textcolor{drd}{its Hessian eigenvectors}, which only reflects some of the relationship between forgetting and landscape. Considering $\boldsymbol{d}$ is randomly selected, we repeat the visualization 10 times with different $\boldsymbol{d}$.

%(refer to Appendix \ref{app_visual} for details). 

\begin{figure}[t]
    \centering
    \includegraphics[width=1.0\textwidth]{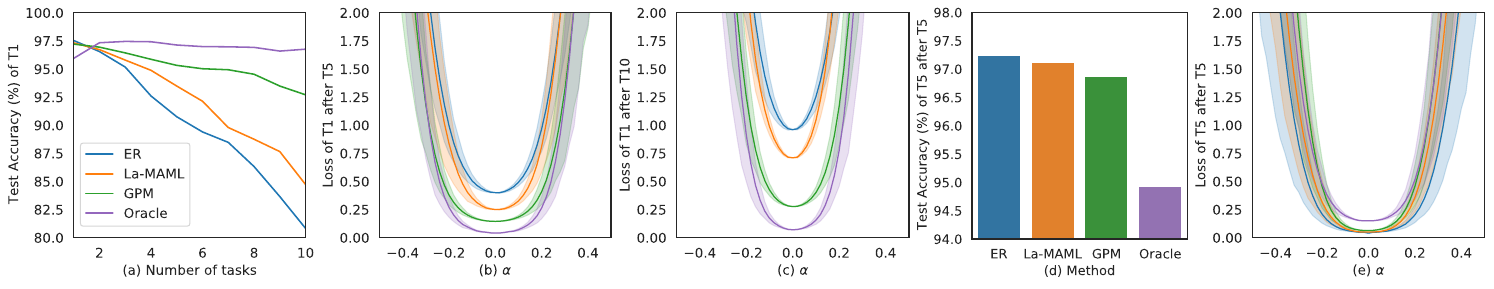}
    \vspace{-5mm}
    %\caption{The connection between the weight loss landscape and the continual learning methods. (a)-(c) Stability of the first task. (a) Test accuracy change curve of the first task; (b) Landscape of first task after learning the fifth task; (c) Landscape of first task after learning all ten tasks. (d)-(e) Sensitivity of the fifth task. (d) Test accuracy and (e) Landscape of the fifth task after just learning the fifth task. ("Landscape" is abbr. of the weight loss landscape. "Task$i$" is abbr. of $i$-th task)}
    \caption{The connection between the weight loss landscape and continual learning is investigated on four methods. (a)-( c) shows the stability of the first task. (a) is the test accuracy change curve of the first task; (b) and (c) are the weight loss landscape of the first task after learning the fifth task and all ten tasks. (d)-(e) shows the sensitivity of the fifth task, which are the test accuracy and the weight loss landscape of the fifth task after just learning the fifth task. The shape of the weight loss landscape obtained using ten different random filter-normalized directions. ("T$i$" is abbr. of the $i$-th task)}
    \vspace{-2mm}
    \label{fig_landscape}
\end{figure}

We first study the \textit{stability} of the network by plotting changes of the weight loss landscape for the first task after new task learning. In particular, We use the previously proposed ER \cite{chaudhry2019continual}, La-MAML \cite{gupta2020maml}, and GPM \cite{saha2021gradient} to train a MLP network with two hidden layers on the Permuted MNIST (PMNIST) \cite{lecun1998gradient} dataset that contains 10 tasks. We also retrain the network on the entire dataset contain all passed tasks as an Oracle network. Early stopping is used to halt the training with up to 10 epochs for each task based on the validation loss as proposed in \cite{serra2018overcoming}. As shown in Figure \ref{fig_landscape}(a), all three continual learning methods lose their stability as learning new tasks. It can be observed from Figure \ref{fig_landscape}(b)-(c) that  the weight loss landscape becomes sharper and loss value increases simultaneously, when the testing accuracy of the first task continually decreases. We further evaluate the \textit{sensitivity} of the network by observing the performance of the fifth task just after it has just been learned. As shown in Figure \ref{fig_landscape}(d)-(e), ER shows the best learning capability compared with other methods with the lowest loss value and the flattest loss landscape. Thus, based on these empirical findings, we assume that lower loss value with a flatter neighbor may lead to better continual learning performance.

\subsection{A Case Study of Flattening Sharpness for Vanilla ER}
\label{subsec_fser}

\begin{wrapfigure}{r}{0.45\textwidth}
\vspace{-0.4cm}
\centering
\includegraphics{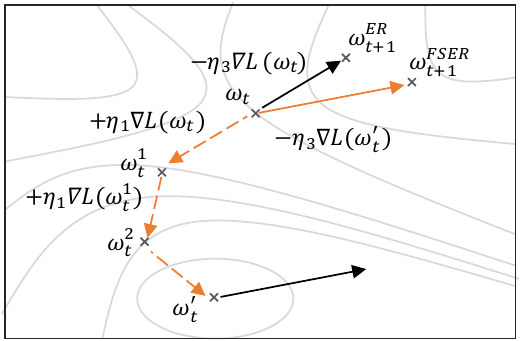}
\vspace{-1mm}
\caption{Schematic of FS-ER update. The dashed line and the solid line indicate the gradient ascent and descent, respectively. Orange denotes the actual update of the parameter $\boldsymbol{w}$.}
\label{fig_gradupdate}
\vspace{2mm}
\includegraphics{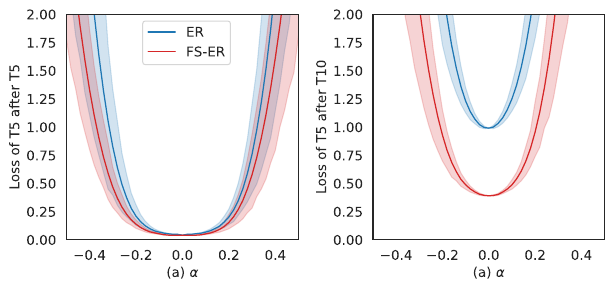}
\vspace{-6mm}
\caption{Landscape of the fifth task after just learning (a) the fifth task and (b) all ten tasks. }
\label{fig_fser}
\vspace{-6mm}
\end{wrapfigure}

% \begin{wrapfigure}{r}{0.48\textwidth}
% \includegraphics[width=0.45\textwidth]{fig_grad_update}
% \caption{Schematic of FS-ER update. The dashed line and the solid line indicate the gradient ascent and descent, respectively. Orange denotes the actual update of the parameter $\boldsymbol{w}$.}
% \label{fig_gradupdate}
% \end{wrapfigure}
% \begin{wrapfigure}{r}{0.48\textwidth}
% \includegraphics[width=0.48\textwidth]{fig_fs_er}
% \caption{Landscape of the fifth task after just learning (a) the fifth task and (b) all ten tasks. }
% \label{fig_fser}
% \end{wrapfigure}

In this part, we further validate our above assumption by Flattening Sharpness for vanilla ER  (FS-ER). Compared with ER that looks for a solution $\boldsymbol{w}$ that jointly minimizes the training loss of current task data and memory data, FS-ER seeks out a solution with both low loss and flat neighbor by minimizing the maximal loss in the neighbor around the parameter value. The schematic of the Flattening Sharpness (FS) is shown in Figure \ref{fig_gradupdate}. We introduce the adversarial weight perturbation (orange dashed line) to explicitly flatten the weight loss landscape via injecting the worst-case weight perturbation, which is calculated from the current task data and past task data sampled from the replay buffer (Refer to Appendix \ref{app_fser_alg} and the next section for more details). Figure \ref{fig_fser} respectively shows the weight loss landscapes of the fifth task after just learning the fifth task and all tasks. We find that FS-ER successfully gets a solution with a lower loss value and flatter landscape, either after the fifth task has just been learned or all ten tasks have been learned. The average testing accuracy of all tasks using FS-ER is $\textbf{90.44\%}$, significantly higher than ER ($\textbf{86.16\%}$), which means that flattening sharpness does benefit continual learning.

\section{Flattening Sharpness for Dynamic Gradient Projection Memory}
\label{sec_method}
% \textcolor{gychen}{Our work is motivated by GPM \cite{saha2021gradient}, which achieves excellent stability on old tasks from theoretical and empirical perspectives. As shown in Figure \ref{fig_landscape}, }
\textcolor{drd}{As shown in Figure \ref{fig_landscape}, GPM achieves the highest testing accuracy on old tasks among all three practical continual learning methods, but shows less sensitivity to new task learning.} To address this issue, we propose \textit{Flattening Sharpness for Dynamic GPM (FS-DGPM)}, which dynamically adjusts the gradient subspace representing the 
past tasks to improve the sensitivity to the new task, while ensuring stability of the previous tasks. In particular, we let $\boldsymbol{M}=[\boldsymbol{u}_1, \boldsymbol{u}_2, \cdots, \boldsymbol{u}_k]$ denote the bases matrix that spans the gradient subspace of the previous task, $\boldsymbol{\Lambda}=\mbox{diag}[\lambda_1, \lambda_2, ..., \lambda_k]$ be the diagonal matrix with its $i$-th diagonal element $\lambda_i\in[0,1]$ indicating the importance of each basis, and $k$ is the number of bases.

\subsection{Sharpness Evaluation} 
Comparing with the classical strategy that perturbs weight in the entire space \cite{wen2018flipout, khan2018fast, wu2020adversarial, foret2020sharpness}, we focus on characterizing the weight loss landscape on the new task with respect to the important subspace representing old task. The important subspace can be effectively calculated based on the examples sampled from replay buffer $\mathcal{M}$ after each task training. Then, we can find the worst case by maximizing the training loss of the network on the new task in this subspace. Formally, the sharpness of the loss landscape around the solution $\boldsymbol{w}$ in the old parameter space can be predicted as,
\begin{equation}
     \max_{\boldsymbol{v} \in \mathcal{V} } \quad  L_{\hat{\mathcal{D}}_t} \left(\boldsymbol{w} + \boldsymbol{v} \right),
     \label{eq_sharpmax}
\end{equation}
where $\mathcal{V}$ denotes the subspace spanned by $\boldsymbol{M}$ and $\boldsymbol{\Lambda}$, and $\hat{\mathcal{D}}_t$ denotes the batch samples of the current $t$-th task. As shown in Eq. (\ref{eq_sharpmax}), the high value can be obtained when the network fails to learn the new task (\textit{sensitivity}) and the new task learning seriously interferes with the past tasks learning (\textit{stability}). Based on the gradient method, the adversarial weight perturbation $\boldsymbol{v}$ can be solved as,
\begin{equation}
\label{eq_sharp}
\boldsymbol{v} \leftarrow \boldsymbol{v} + \eta_{1} \boldsymbol{M} \boldsymbol{\Lambda} \boldsymbol{M}^{T} \left(\nabla_{(\boldsymbol{w}+\boldsymbol{v})} L_{\hat{\mathcal{D}}_t}(\boldsymbol{w}+\boldsymbol{v})\right),
%\boldsymbol{v} \leftarrow \boldsymbol{v}+\eta_{1} \cdot \Pi_{\mathbb{S}}\left( \nabla_{ (\boldsymbol{w}+\boldsymbol{v})} L_{\hat{\mathcal{D}}}(\boldsymbol{w}+\boldsymbol{v}) \right),
\end{equation}
where $\eta_{1}$ is the update step size. Note that $\boldsymbol{v}$ is initialized as $\mathbf{0}$ and layer-wise updated. As shown in Appendix \ref{app_exp_result}, two-step for $\boldsymbol{v}$ (default settings) are enough to get good improvements.

\subsection{Dynamic Gradient Projection Memory}
After obtaining the adversarial weight perturbation $\boldsymbol{v}$, we can further update the bases importance matrix $\boldsymbol{\Lambda}=\mbox{diag}[\lambda_1, \lambda_2, ..., \lambda_k]$ by jointly considering the current task batch $\hat{\mathcal{D}}_t$ and the batch $\hat{\mathcal{M}}$ sampled from the replay buffer $\mathcal{M}$ as following,
\begin{equation}
\label{eq_lam}
\lambda_i \leftarrow \lambda_i - \eta_{2} \left(\nabla_{\lambda_i} L_{\hat{\mathcal{D}}_t \cup \hat{\mathcal{M}}}(\boldsymbol{w}+\boldsymbol{v}) \right),
\end{equation}
where the sigmoid function is used \textcolor{drd}{at the end of gradient update} to constrain the importance value $\lambda_i$ between 0 and 1. In addition, the second term on the right side \textcolor{drd}{in Eq. (\ref{eq_lam})} can be approximated by the first-order Taylor expansion as,
\begin{equation}
\nabla_{\lambda_i} L_{\hat{\mathcal{D}}_t \cup \hat{\mathcal{M}}}(\boldsymbol{w}+\boldsymbol{v})
        \approx \eta_{1} \left( \nabla_{\boldsymbol{w}} L_{\hat{\mathcal{D}}_t}\left(\boldsymbol{w}\right)\right)^T \boldsymbol{u}_{i} \boldsymbol{u}_{i}^{T} \left( \nabla_{\boldsymbol{w}} L_{\hat{\mathcal{D}}_t \cup \hat{\mathcal{M}}}\left(\boldsymbol{w}\right) \right).
\end{equation}

\begin{wrapfigure}{r}{0.48\textwidth}
\centering
\includegraphics[width=0.48\textwidth]{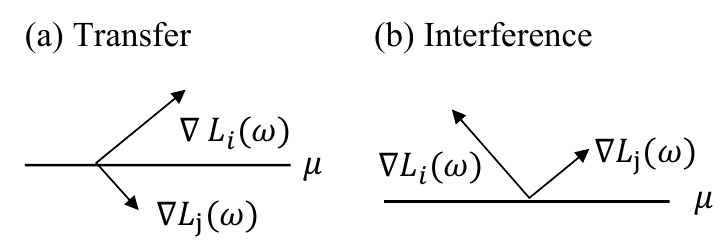}
\vspace{-5mm}
\caption{A depiction of transfer (a) and interference (b) in the basis $\boldsymbol{u}$ of gradient space.}
\label{fig_grad_relation}
\end{wrapfigure}
The above equation characterizes the relationship between the gradients induced by the current task and the old tasks based on the basis $\boldsymbol{u}_{i}$. As illustrated in the Figure \ref{fig_grad_relation}, this equation implies that when the projections of two gradients on the basis $\boldsymbol{u}_{i}$ are aligned in the same direction, the gradient of $\lambda_i$ will be positive, and when there is interference, the gradient will be negative. The positive (negative) gradient will decrease (increase) the importance $\lambda_i$, thereby releasing (tightening) the update limit of the new task on the corresponding basis $\boldsymbol{u}_{i}$. We provide the full derivation in the Appendix \ref{app_dev_weight}. \textcolor{drd}{Note that the initial value of all importance is set to 1 and dynamically adjusted from the second task.}
% Note that the GPM is dynamically adjusted from the second task.

\subsection{Weight Updating}

Finally, we update the model parameters by minimizing the worst performance of $f_{\boldsymbol{w}+\boldsymbol{v}}$ with respect to $\boldsymbol{w}$, while adjusting the update magnitude of $\boldsymbol{w}$ on each basis based on its importance to alleviate forgetting. More concretely, the parameter $\boldsymbol{w}$ will be updated to:
\begin{equation}
\label{eq_weight}
\boldsymbol{w} \leftarrow \boldsymbol{w} -\eta_3 \left( \boldsymbol{I} - \boldsymbol{M} \boldsymbol{\Lambda} \boldsymbol{M}^{T} \right) \nabla_{\boldsymbol{w}} L_{\hat{\mathcal{D}}_t \cup \hat{\mathcal{M}}}(\boldsymbol{w}+\boldsymbol{v}).
\end{equation}
Note that the optimization is performed over the model parameters $\boldsymbol{w}$, whereas the objective is computed using perturbed model $f_{\boldsymbol{w}+\boldsymbol{v}}$. In addition, we update the replay buffer $\mathcal{M}$ with reservoir sampling as in \cite{riemer2018learning}, and then use Singular Value Decomposition (SVD) to recalculate $\boldsymbol{M}$ based on the sampling data in the replay buffer after learning one task following GPM  \cite{saha2021gradient}. Comparing with \cite{saha2021gradient}, we calculate the important bases in the entire gradient space and use them to replace the bases calculated last time. \textcolor{drd}{Besides that, our method degenerates to GPM when $\eta_1$ and $\eta_2$ are set to $0$.} The complete pseudo-code of FS-DGPM is outlined in the Algorithm \ref{alg-fsdgpm}.% \textcolor{gychen}{I think we can include algorithm in our main paper if there is enough space left}.

\begin{algorithm}[H]
	\caption{FS-DGPM (Flattening Sharpness for Dynamic Gradient Projection Memory)} 
	\label{alg-fsdgpm}
	\begin{algorithmic}
		\State  \textbf{Input:} Network weight $\boldsymbol{w}$, loss function $\ell$, learning rate $\eta_3$, FS step size $\eta_1$, FS steps $K$, Soft weight step size $\eta_2$, batch size $b$.
		\State  Initializing $\mathcal{M} \leftarrow\{\} $,  $\boldsymbol{M} \leftarrow \boldsymbol{I},  \boldsymbol{\Lambda} \leftarrow \boldsymbol{I}$ 
		\For{$t=1,2,\cdots,T$}
		\For{$ep=1,2,\cdots,num_{epochs}$}
		\For{batch $\hat{\mathcal{D}}_t \stackrel{b}{\sim} \mathcal{D}_t$}
		\State $\hat{\mathcal{M}} \stackrel{b}{\sim} \mathcal{M}$
% 		\For{batch $\hat{\mathcal{D}}_t \subseteq \mathcal{D}_t$}
% 		\State $\hat{\mathcal{M}} \leftarrow$ Sample $(\mathcal{M})$
		\For{$k=1,\cdots,K$}
		\State $\boldsymbol{v} \leftarrow \boldsymbol{v} + \eta_{1} \boldsymbol{M} \boldsymbol{\Lambda} \boldsymbol{M}^{T} \left(\nabla_{(\boldsymbol{w}+\boldsymbol{v})} L_{\hat{\mathcal{D}}}(\boldsymbol{w}+\boldsymbol{v})\right)$ {\color{gray!66} \Comment{Sharpness Evaluation}}
		\EndFor
		\If{$t \geq 2$}
		\State $\boldsymbol{\Lambda} \leftarrow \boldsymbol{\Lambda} - \eta_{2} \left(\nabla_{\boldsymbol{\Lambda}} L_{\hat{\mathcal{D}}_t \cup \hat{\mathcal{M}}}(\boldsymbol{w}+\boldsymbol{v}) \right)$ {\color{gray!66} \Comment{Dynamic Gradient Projection Memory}}
		\EndIf
		\State $\boldsymbol{w} \leftarrow \boldsymbol{w} -\eta_3 \left( \boldsymbol{I} - \boldsymbol{M} \boldsymbol{\Lambda} \boldsymbol{M}^{T} \right) \nabla_{\boldsymbol{w}} L_{\hat{\mathcal{D}}_t \cup \hat{\mathcal{M}}}(\boldsymbol{w}+\boldsymbol{v})$ {\color{gray!66} \Comment{Weight updating}}
		\State Push $\hat{\mathcal{D}}_t$ to $\mathcal{M}$ with reservior sampling
		\EndFor
		\EndFor
		\State $\boldsymbol{M} \leftarrow$ UpdateGPM $(\mathcal{M})$ {\color{gray!66} \Comment{see Appendix Alg. \ref{alg_base}}}
		\EndFor
	\end{algorithmic}
\end{algorithm}

\subsection{Theoretical Understanding}
We further provide a theoretical view on why landscape can characterize the continual learning performance and why our proposed FS-DGPM works. To simplify our explanation, we only consider two tasks, which contains the training sets $\mathcal{D}_{1}$ and $\mathcal{D}_{2}$ sampled from the distributions $\mathscr{D}_{1}$ and $\mathscr{D}_{2}$, respectively. Based on the previous works on PAC-Bayes bound \cite{neyshabur2017exploring,wu2020adversarial, foret2020sharpness}, given a "prior" distribution $P$ (a common assumption is zero mean, $\sigma^2$ variance Gaussian distribution) over the weights, the expected error of the classifier for the continual learning scenario can be bounded with probability at least $1-\delta$ over the draw of $n$ training data:
\begin{equation}
\textcolor{drd}{
\begin{aligned}
&\min_{\Delta \boldsymbol{w}} \mathbb{E}_{\boldsymbol{v}}\left[L_{\mathscr{D}_{1} \cup \mathscr{D}_{2}}(\boldsymbol{w}+\Delta \boldsymbol{w} +\boldsymbol{v}) \right] \leq \min_{\Delta \boldsymbol{w}\in  \mathcal{V}^C} \mathbb{E}_{\boldsymbol{v}}\left[L_{\mathscr{D}_{1}}(\boldsymbol{w} +\Delta \boldsymbol{w} +\boldsymbol{v})\right] + L_{\mathcal{D}_{2}}(\boldsymbol{w}+\Delta \boldsymbol{w}) \\
&+\max_{\boldsymbol{v}\in \mathcal{V}} L_{\mathcal{D}_{2}}(\boldsymbol{w}+\Delta \boldsymbol{w}+\boldsymbol{v}) -L_{\mathcal{D}_{2}}(\boldsymbol{w}+\Delta \boldsymbol{w}) 
+ 4\sqrt{\frac{1}{n}\left(KL(\boldsymbol{w}+\Delta \boldsymbol{w} +\boldsymbol{v}||P)+\ln\frac{2n}{\delta}\right)}.
\end{aligned}
}\nonumber
\end{equation}

where $\Delta \boldsymbol{w}$ is the update based on the previously optimal solution $\boldsymbol{w}$ learned on the old task $\mathcal{D}_{1}$ when learning the new one $\mathcal{D}_{2}$, and $\boldsymbol{v}$ is often chosen as a zero mean spherical Gaussian perturbation with variance $\sigma^2$ in every direction. Let $\Delta \boldsymbol{w}\in  \mathcal{V}^C$, then $\Delta \boldsymbol{w}$ lies in the complementary space of the important space representing the old task $\mathscr{D}_1$, so that  $\mathbb{E}_{\boldsymbol{v}}\left[L_{\mathscr{D}_{1}}(\boldsymbol{w} +\Delta \boldsymbol{w} +\boldsymbol{v})\right]$ does not increase too much compared with the previously minimized $\mathbb{E}_{\boldsymbol{v}}\left[L_{\mathscr{D}_{1}}(\boldsymbol{w}+\boldsymbol{v})\right]$. The second term denotes the empirical loss on the second task and the third term represents the sharpness of the weight loss landscape around the $\boldsymbol{w}+\Delta\boldsymbol{w}$. \textcolor{drd}{Since we have constrained $\Delta\boldsymbol{w}\in\mathcal{V}^C$, then it is natural to assume $\boldsymbol{v}\in\mathcal{V}$, so that $\Delta \boldsymbol{w}+\boldsymbol{v}$ will cover the full space.} Thus, our FS-DGPM exactly optimizes the worst-case of the flatness of weight loss landscape to control the PAC-Bayes bound, which theoretically justifies both lower loss value and flatter landscape lead to better continual learning performance, and why our proposed FS-DGPM works.

\section{Experiments}
\label{exp}

In this section, we conduct extensive experiments to compare the performance of our proposed FS-DGPM model with the state-of-the-art methods on widely used continual learning benchmark datasets. Additional results and more details about the datasets, experiment setup, baselines, and model architectures are presented in the Appendix \ref{app_exp} and \ref{app_exp_result}.

\subsection{Experimental Setup}
\label{setup}

\paragraph{Datasets:} 
We evaluate our algorithm on four image classification datasets: \textbf{Permuted MNIST} (PMNIST) \cite{lecun1998gradient}, \textbf{CIFAR-100 Split} \cite{krizhevsky2009learning}, \textbf{CIFAR-100 Superclass} \cite{yoon2019scalable} and \textbf{TinyImageNet} \cite{tinyImagenet}. The PMNIST dataset is a variant of the MNIST dataset, in which each task applies a fixed random pixel permutation to the original dataset. The PMNIST benchmark dataset consists of 20 tasks, and each contains only 1000 samples from 10 different classes \cite{gupta2020maml}. The CIFAR-100 Split is constructed by randomly dividing 100 classes of CIFAR-100 into 10 tasks with 10 classes per task. The CIFAR-100 Superclass is divided into 20 tasks according to the 20 superclasses of the CIFAR-100 dataset, and each superclass contains 5 different but semantically related classes. Whereas, TinyImageNet is constructed by splitting 200 classes into 40 5-way classification tasks. In our experiments, we do not use any data augmentation. The dataset statistics are given in Appendix \ref{app_dataset}.

\paragraph{Network Architecture:}
For PMNIST, we use a fully connected network with two hidden layers of 100 units each following \cite{lopez2017gradient}. For experiments of CIFAR-100 Split and CIFAR-100 Superclass, we use a 5-layer AlexNet and LeNet architecture similar to \cite {saha2021gradient} respectively. For TinyImageNet, we use the same network architecture as \cite{gupta2020maml}, which consists of 4 conv layers and 3 fully connected layers. In PMNIST, all tasks share the final classifier layer, while other experiments use a multi-head incremental setting, that is, each task has a separate classifier. Refer to Appendix \ref{app_arch} for more details.

\paragraph{Baselines:}
We compare our method against multiple methods described below. (1) \textbf{EWC} \cite{kirkpatrick2017overcoming}, a regularization-based method that uses the diagonal of Fisher information to identify important weights; (2) \textbf{ICARL} \cite{rebuffi2017icarl}, a memory-based method that uses knowledge-distillation and episodic memory to reduce forgetting; (3) \textbf{GEM} \cite{lopez2017gradient}, another memory-based method that uses the gradient of episodic memory to constrain optimization to prevent forgetting; (4) \textbf{ER} \cite{chaudhry2019continual}, a simple and competitive method based on reservoir sampling; (5) \textbf{La-MAML} \cite{gupta2020maml} and (6) \textbf{GPM} \cite{saha2021gradient} are memory-based methods inspired by optimization-based meta-learning and based on gradient orthogonal constraints, respectively; (7) \textbf{Multitask} is an oracle baseline that all tasks are learned jointly using the entire dataset at once in a single network. Multitask is not a continual learning strategy but serves as an upper bound on average test accuracy on all tasks. 

\paragraph{Training Details:}
All baselines and our method use stochastic gradient descent (SGD) for training. For each task in PMNIST and TinyImageNet, we train the network in 1 and 10 epochs, respectively, with the batch size as 10. These experimental settings are the same as La-MAML \cite {gupta2020maml}, so that we directly compare with their reported results. In the CIFAR-100 Split and CIFAR-100 Superclass experiments, we use the early termination strategy to train up to 50 epochs for each task, which is based on the validation loss as proposed in \cite{serra2018overcoming}. For both datasets, the batch size is set to 64. The replay buffer size of PMNIST, CIFAR-100 Split, CIFAR-100 Superclass, and TinyImageNet are 200, 1000, 1000, and 400, respectively. Details about the experimental setting and the hyperparameters considered for each baseline are provided in Appendix \ref{app_thres} and \ref{app_hyperpara}.

\paragraph{Metrics:}
We evaluate the continual learning performance by the \textit{average accuracy} (ACC) and \textit{backward transfer} (BWT) \cite{lopez2017gradient, chaudhry2018efficient, chaudhry2019continual}, formulated as following,
\begin{equation*}
ACC = \frac{1}{T} \sum_{i=1}^{T} R_{T,i} , \quad BWT = \frac{1}{T-1} \sum_{i=1}^{T-1} R_{T,i} - R_{i,i},
\end{equation*}
where $T$ is the total number of learned sequential tasks, $R_{i, j}$ is the test classification accuracy of the model on $j$-th task after learning the last sample from $i$-th task. ACC is the average test classification accuracy of all tasks, bigger is better. BWT is the interference of new learning on the past knowledge. More specifically, negative BWT implies (catastrophic) forgetting whereas positive BWT indicates learning new task increases the performance on some preceding tasks. 

\subsection{Results and Discussion}
\label{results}
\begin{figure}[t]
    \centering
    \includegraphics[width=1.0\textwidth]{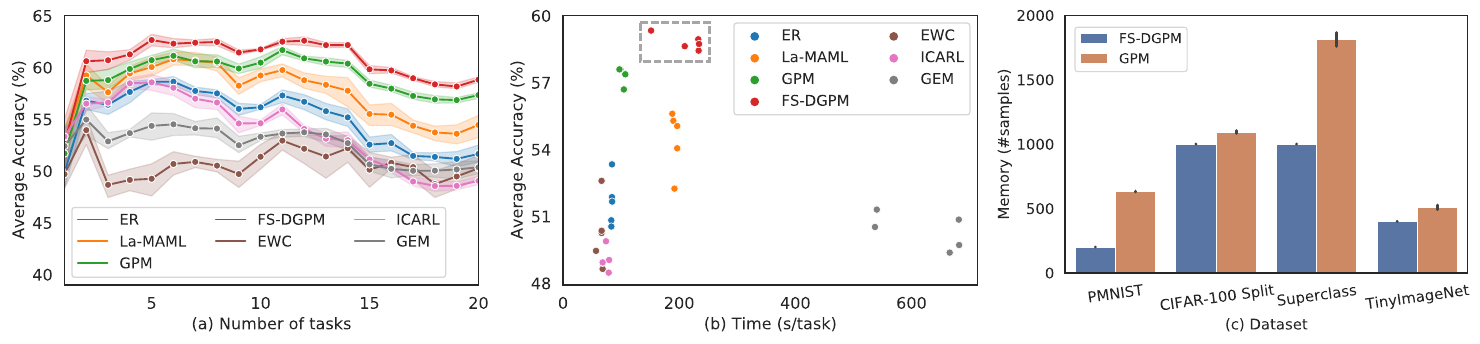}
    %\vspace{-2mm}
    \caption{(a) Average accuracy as a function of the number of tasks trained on 20-Split CIFAR-100 Superclass. (b) Training time per task on 20-Split CIFAR-100 Superclass. (c) Memory usage on four datasets. ("Superclass" is abbr. of CIFAR-100 Superclass).}
    \label{fig_exp}
\end{figure}
\begin{table}[t]
	\caption{Experimental results on 10-Split CIFAR-100, 20-Split CIFAR-100 Superclass and 40-Split TinyImageNet in 50 epochs. Each experiment is run with 5 seeds. $^\dag$ and $^*$ denotes results reported by \cite{gupta2020maml} and \cite{yoon2019scalable} respectively. }
	\label{tab_cifar}
	\begin{center}
	\resizebox{\textwidth}{!}{
		\begin{tabular}{lcccccc}
			\toprule
			&\multicolumn{2}{c}{\textbf{CIFAR-100 Split}}  &\multicolumn{2}{c}{\textbf{CIFAR-100 Superclass}} &\multicolumn{2}{c}{\textbf{TinyImageNet}} \\ 
			\cmidrule(lr){2-3}  \cmidrule(lr){4-5} \cmidrule(lr){6-7} 
			\textbf{Method}     &ACC(\%)      &BWT(\%)     &ACC(\%)      &BWT(\%)     &ACC(\%)      &BWT(\%) \\ \midrule
			EWC       & 72.77 $\pm$ 0.45  & -3.59 $\pm$ 0.55  & 50.26 $\pm$ 1.48  & -7.87 $\pm$ 1.63 & -  & - \\
			GEM     & 70.15 $\pm$ 0.34  & -8.61 $\pm$ 0.42 & 50.35 $\pm$ 0.80  & -9.50 $\pm$ 0.85 & 50.57 $\pm$ 0.61$^\dag$  & -20.50 $\pm$ 0.10$^\dag$ \\
			%A-GEM     	 & 65.90 $\pm$ 0.71  & -12.28 $\pm$ 0.82  & 49.68 $\pm$ 1.09  & -8.55 $\pm$ 1.12 & 46.38 $\pm$ 1.34  & -19.96 $\pm$ 0.61 \\
			ICARL & 53.50 $\pm$ 0.81 & -20.44 $\pm$ 0.82 & 49.05 $\pm$ 0.51  & -11.24 $\pm$ 0.27 & 54.77 $\pm$ 0.32$^\dag$  & -3.93 $\pm$ 0.55$^\dag$ \\
			ER   & 70.07 $\pm$ 0.35 & -7.70 $\pm$ 0.59 & 51.64 $\pm$ 1.09  & -7.86 $\pm$ 0.89 & 48.32 $\pm$ 1.51$^\dag$  & -19.86 $\pm$ 0.70$^\dag$ \\
			La-MAML    & 71.37 $\pm$ 0.67 &  -5.39 $\pm$ 0.53 & 54.44 $\pm$ 1.36  & -6.65 $\pm$ 0.85 & 66.90 $\pm$ 1.65$^\dag$  & -9.13 $\pm$ 0.90$^\dag$ \\
			GPM  & 73.18 $\pm$ 0.52 & \textbf{-1.17} $\pm$ \textbf{0.27} & 57.33 $\pm$ 0.37  & \textbf{-0.37} $\pm$ \textbf{0.12} & 67.39 $\pm$ 0.47  & \textbf{1.45} $\pm$ \textbf{0.22} \\
			%\midrule
			\textbf{FS-DGPM}  & \textbf{74.33} $\pm$ \textbf{0.31}  & -2.71 $\pm$ 0.17 & \textbf{58.81} $\pm$ \textbf{0.34}  & -2.97 $\pm$ 0.35 & \textbf{70.41} $\pm$ \textbf{1.30}  & -2.11 $\pm$ 0.84 \\
			\midrule
			Multitask & 79.75 $\pm$ 0.38 & - & 61.00 $\pm$ 0.20$^*$ & - & 77.10 $\pm$ 1.06$^\dag$ & - \\
			\bottomrule
		\end{tabular}}
	\end{center}
\end{table}

\begin{wraptable}{r}{0.5\textwidth}
\vspace{-10mm}
\setlength{\abovecaptionskip}{-0.2cm}
\setlength{\belowcaptionskip}{-0.cm}
	\caption{Experimental results (mean $\pm$ std in 5 runs) on PMNIST in single-epoch.}
	\label{tab_pmnist}
	\begin{center}
% 	\resizebox{0.5\textwidth}{!}{
		\begin{tabular}{lcccc}
			\toprule
			    &\multicolumn{2}{c}{\textbf{PMNIST}}  \\ 
			\cmidrule(lr){2-3} 
			\textbf{Method}      &ACC(\%)      &BWT(\%)      \\ \midrule
			EWC           & 62.25 $\pm$ 1.44 & -15.22 $\pm$ 1.25 \\
			GEM     	 & 61.82 $\pm$ 0.85 & -15.58 $\pm$ 1.17 \\
			ER        	 & 68.31 $\pm$ 0.56 & -13.91 $\pm$ 0.67 \\
			La-MAML      & 75.98 $\pm$ 0.60 & -10.21 $\pm$ 0.90 \\
			GPM          & 74.54 $\pm$ 0.36 & \textbf{-7.17} $\pm$ \textbf{0.51} \\
			\textbf{FS-DGPM}  & \textbf{76.96} $\pm$ \textbf{0.51} & -7.45 $\pm$ 0.30 \\
			\midrule
			Multitask & 86.54 $\pm$ 1.74 & - \\
			\bottomrule
		\end{tabular}
% 		}
	\end{center}
	\vspace{-5mm}
\end{wraptable}
\paragraph{PMNIST:}
We first evaluate our algorithm for 20 sequential PMNIST tasks with only 1000 samples per task in a single-head setting. From the results, as shown in Table \ref{tab_pmnist}, we see that our method (FS-DGPM) achieves the best average accuracy $(\textbf{76.96\%} \pm \textbf{0.77})$. Moreover, we achieve the least amount of forgetting except GPM, which is essentially a trade-off in accuracy to minimize forgetting. As shown in Figure \ref{fig_exp}(c), we only use about $\textbf{31\%}$ of the final memory size of GPM and achieve $\sim \textbf{2.5\%}$ better accuracy.

\paragraph{CIFAR-100 and TinyImageNet:} Next, we use a multi-head setting to evaluate our algorithm under the more challenging visual classification benchmarks. Table \ref{tab_cifar} reports all results of these experiments. We outperform all baselines on three datasets, with achieving the best average accuracy $\textbf{74.33\%}$, $\textbf{58.81\%}$ and $\textbf{70.41\%}$. In these experiments, we observe that GPM is a strong baseline with the least forgetting. At the same time, we highlight that our method achieves the highest accuracy on all datasets and the second-lowest forgetting after GPM. Figure \ref{fig_exp}(a) shows the process of performance changing with the number of tasks on the CIFAR-100 Superclass. We consistently see the superior performance of our method at any stage of model evolution.
It is also worth emphasizing that although our method requires more time for training than GPM, it has lower memory usage and better test accuracy (See Figure \ref{fig_exp}(b)-(c)). As noted by \cite{chaudhry2018efficient} and \cite{tang2020graph}, EWC performs poorly without multiple passes over the datasets, and GEM is not very effective under the single-head variants. These situations have also been observed in our experiments.

\begin{figure}[t]
    \centering
    \includegraphics[width=1.0\textwidth]{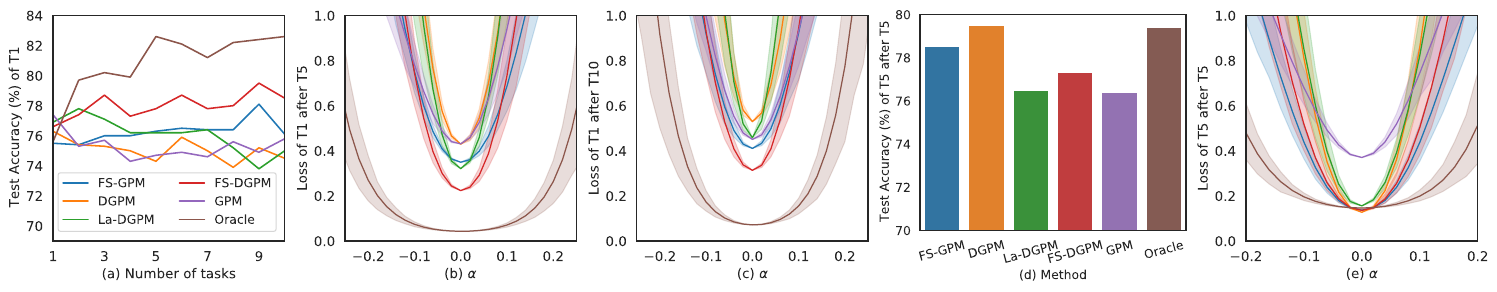}
    % \vspace{-3mm}
    %\caption{The ablation study experiments on CIFAR-100 Split with 10 tasks. (a) Test accuracy change curve of the first task; (b) Landscape of the first task after learning the fifth task; (c) Landscape of first task after learning all ten tasks; (d) Test accuracy of the fifth task after just learning the fifth task; (e) Landscape of the fifth task after just learning the fifth task.}
    \caption{The ablation study implemented on CIFAR-100 Split with 10 tasks. (a)-(c) shows the stability of the first task. (a) is the test accuracy change curve of the first task; (b) and (c) are the weight loss landscape of the first task after learning the fifth task and all ten tasks. (d)-(e) shows the sensitivity of the fifth task, which are the test accuracy and the weight loss landscape of the fifth task after just learning the fifth task. The shape of the weight loss landscape obtained using ten different random filter-normalized directions. ("T$i$" is abbr. of the $i$-th task)}
    \label{fig_exp_ablation}
    % \vspace{-3mm}
\end{figure}

\subsection{Ablation studies on FS-DGPM}
\begin{wraptable}{r}{0.6\textwidth}
\setlength{\abovecaptionskip}{-0.25cm}
\setlength{\belowcaptionskip}{-0.cm}
	\caption{The ablation study results on CIFAR-100 Split and Superclass. Each experiment is run with 5 seeds.}
	\label{tab_ablation}
	\begin{center}
	\resizebox{0.6\textwidth}{!}{
		\begin{tabular}{lcccc}
			\toprule
			    &\multicolumn{2}{c}{\textbf{CIFAR-100 Split}}  &\multicolumn{2}{c}{\textbf{CIFAR-100 Superclass}} \\ 
			\cmidrule(lr){2-3}  \cmidrule(lr){4-5}
			\textbf{Method}  &ACC(\%)  &BWT(\%)   &ACC(\%)  &BWT(\%)     \\ \midrule
			FS-DGPM   & \textbf{74.33} $\pm$ 0.31 & \textbf{-2.71} $\pm$ 0.17 & \textbf{58.81} $\pm$ 0.34 & -2.97 $\pm$ 0.35 \\
			La-DGPM   & 73.74 $\pm$ 0.61 & -3.05 $\pm$ 0.73 & 58.18 $\pm$ 0.41 & \textbf{-2.41} $\pm$ 0.39 \\
			FS-GPM   & 73.96 $\pm$ 0.44 & -3.12 $\pm$ 0.43 & 58.61 $\pm$ 0.53 & -2.79 $\pm$ 0.30 \\
			DGPM     & 73.78 $\pm$ 0.32 & -3.67 $\pm$ 0.42 & 56.78 $\pm$ 0.49 & -2.44 $\pm$ 0.40 \\
			\midrule
			GPM & 73.18 $\pm$ 0.52 & \textbf{-1.17} $\pm$ 0.27 & 57.33 $\pm$ 0.37 & \textbf{-0.37} $\pm$ 0.12 \\
			\bottomrule
		\end{tabular}}
	\end{center}
	\vspace{-2mm}
\end{wraptable}
We further investigate our model performance with an ablation study and summarize it in Table \ref{tab_ablation}. We respectively ablate the effects of flattening sharpness and dynamically adjusting the soft weight for bases. We refer to them as DGPM and FS-GPM. We also construct an ablation referred to as La-DGPM (Look-ahead DGPM), where the adversarial weight perturbation $\boldsymbol{v}$ in Eq. (\ref{eq_sharp}) is changed to the direction of gradient descent. At the same time, we also change the sign in Eq. (\ref{eq_lam}) to ensure that the soft weight of the basis is reduced when the gradients are aligned in the same direction. From the results, shown in Table \ref{tab_ablation}, we observe that flattening sharpness does benefits GPM, with $\sim \textbf{1.0\%}$ improvement over GPM on both datasets. We can further observe through Figure \ref{fig_exp_ablation} that all landscapes of FS-DGPM have lower loss values and flatter neighbors than DGPM and La-DGPM on the CIFAR-100 Split experiments. In addition, we see that DGPM performs well in learning new tasks, but it also leads to forgetting previous tasks. This situation can be efficiently alleviated by flattening sharpness. Hence, FS is indeed a powerful method worthy of being widely adopted for continual learning scenarios.

\section{Conclusion}
\label{sec_conclu} 

In this paper, we explore the weight loss landscape to characterize the well-known 'sensitivity-stability' dilemma faced by continual learning algorithms, and find that lower loss value with flatter neighbor often leads to better continual learning performance. Based on this finding, we propose our FS-DGPM algorithm, which introduces a soft weight to represent the importance of each basis representing past tasks in GPM, so that less important bases can be dynamically released to improve the sensitivity of new skill learning. Flattening Sharpness (FS) is also introduced here to reduce the generalization gap by explicitly regulating the flatness of the weight loss landscape of all tasks seen so far. The evaluation of various image classification tasks with different network architectures and the comparison with some state-of-the-art algorithms show the effectiveness of our method in achieving high classification performance while alleviating forgetting. Although our method theoretically and empirically demonstrates the advantages of introducing FS into continual learning, whether there exists a closer upper bound for the continual learning performance remains an unresolved problem and left for our future exploration. \textcolor{drd}{Although our method theoretically and empirically demonstrates the advantages of introducing bases soft weight and FS into continual learning, whether there exists a better dynamic adjustment and a closer upper bound for the continual learning performance remains an unresolved problem and left for our future exploration.}

% Although our method theoretically and empirically demonstrates the advantages of introducing FS into continual learning, whether there exists a closer upper bound for the continual learning performance remains an unresolved problem and left for our future exploration.

%  gap with the PAC-Bayes bound. %Moreover, whether there is a method of decoupling low loss and flatten landscape to optimize the performance of continual learning is still an unresolved problem. 
%In this paper, we first investigate sensitivity-stability from the perspective of the weight loss landscape. Based on the observations, we propose a novel continual learning algorithm, FS-DGPM, which improves the sensitivity of learning new skills while maintaining stability by explicitly regulating the flatness of the weight loss landscape of previous tasks. The evaluation of various image classification tasks with different network architectures and the comparison with some state-of-the-art algorithms show the effectiveness of our method in achieving high classification performance while reducing forgetting. 

\section*{Acknowledgments}
\label{sec_ack} 
\textcolor{drd}{We would like to thank Dr. Yongzhe Deng and Zhuo Zhang for their helpful discussions, and anonymous reviewers for their valuable comments to improve this work.} This work was supported by a grant from the National Key Research and Development Program of China (Project No. 2020YFB1313900), Hong Kong Research Grants Council under General Research Fund (Project No. 14201620),  the National Natural Science Foundation of China (Project No. 62006219, 62072452) and Guangdong Provincial Basic and Applied Basic Research Fund-Regional Joint Fund (Project No. 2020B1515130004).
%The work is supported by the Key-Area Research and Development Program of Guangdong Province, China (2020B010165004) and the National Natural Science Foundation of China (Grant Nos.: 62006219, U1813204).

%\clearpage
\bibliography{FS_DGPM}
\bibliographystyle{plain}
%\bibliographystyle{unsrtnat}

%%%%%%%%%%%%%%%%%%%%%%%%%%%%%%%%%%%%%%%%%%%%%%%%%%%%%%%%%%%%

% \clearpage
% \input{checklist.tex}

%%%%%%%%%%%%%%%%%%%%%%%%%%%%%%%%%%%%%%%%%%%%%%%%%%%%%%%%%%%%

\clearpage

\appendix
\section{Appendix}

\subsection{PAC Bayesian Bound}
\label{app_pac_bound}

In this part, we provide a detailed PAC-Bound based on the continual learning scenario.

Based on the previous works on PAC-Bayes bound \cite{neyshabur2017exploring,wu2020adversarial,foret2020sharpness}, let $\ell(\cdot, \cdot)$ be 0-1 loss, then for the empirical loss over the training set $\mathcal{D} \sim \mathscr{D}$, we have $L_{\mathcal{D}}(\boldsymbol{w}) = \frac{1}{|\mathcal{D}|}\sum_{\left(\boldsymbol{x}, y\right)\in \mathcal{D}}\left[\ell \left(f_{\boldsymbol{w}}\left(\boldsymbol{x} \right), y\right)\right] \in [0, 1]$. Given a "prior" distribution $P$ (a common assumption is zero mean, $\sigma^2$ variance Gaussian distribution) over the weights of the form $\boldsymbol{w}+\boldsymbol{v}$, the expected error of $f_{\boldsymbol{w}+\boldsymbol{v}}$ can be bounded with probability at least $1-\delta$ over the draw of $n$ training data:
\begin{equation}
\begin{aligned}
\label{eq_pac1}
\mathbb{E}_{\boldsymbol{v}}\left[L_{\mathscr{D}}(\boldsymbol{w}+\boldsymbol{v})\right] \leq L_{\mathcal{D}}(\boldsymbol{w}) + \mathbb{E}_{\boldsymbol{v}}\left[L_{\mathcal{D}}(\boldsymbol{w} + \boldsymbol{v})\right] - L_{\mathcal{D}}(\boldsymbol{w}) + 4\sqrt{\frac{1}{n}\left( KL(\boldsymbol{w}+\boldsymbol{v}||P)+\ln\frac{2n}{\delta}\right)},
\end{aligned}
\end{equation}
where $\boldsymbol{w}$ is the weight of the predictor learned from the training set, $\boldsymbol{v}$ is a random variable and is often chosen as a zero mean spherical Gaussian perturbation with variance $\sigma^2$ in every direction. 

We now consider the bound in the continual learning scenario. To simplify our explanation, we only consider two tasks, which contains the training sets $\mathcal{D}_{1}$ and $\mathcal{D}_{2}$ sampled from the distributions $\mathscr{D}_{1}$ and $\mathscr{D}_{2}$, respectively. We assume the model $f_{\boldsymbol{w}}$ is learned from the training set $\mathcal{D}_{1}$ ($\boldsymbol{w} = \mathrm{argmin}_{\boldsymbol{w}} L_{\mathcal{D}_{1}}(\boldsymbol{w}$), and then continue to learn the training set $\mathcal{D}_{2}$. Our final goal is to find an optimal parameter $\Delta \boldsymbol{w}$ to minimizes the overall risk $L_{\mathscr{D}_{1} \cup \mathscr{D}_{2}}(\boldsymbol{w}+\Delta \boldsymbol{w})$ for all tasks as follows:
$$
\min_{\Delta \boldsymbol{w}} L_{\mathscr{D}_{1} \cup \mathscr{D}_{2}}(\boldsymbol{w}+\Delta \boldsymbol{w}) $$
% \begin{equation}
% \begin{aligned}
% \text{we have} \ \ \mathcal{D}_{1} \sim \mathscr{D}_{1}, \mathcal{D}_{2} \sim \mathscr{D}_{2}, \boldsymbol{w} = \underset{\boldsymbol{w}}{\mathrm{argmin}} L_{\mathcal{D}_{1}}(\boldsymbol{w})\\
% \min_{\Delta \boldsymbol{w}} \quad L_{\mathscr{D}_{1} \cup \mathscr{D}_{2}}(\boldsymbol{w}+\Delta \boldsymbol{w})
% \end{aligned}
% \end{equation}
Based on Eq. (\ref{eq_pac1}), the expected error of $f_{\boldsymbol{w}+\Delta \boldsymbol{w}+\boldsymbol{v}}$ can be bounded with probability at least $1-\delta$ over the draw of training set $\mathcal{D}_{2}$:
\begin{equation}
\begin{aligned}
&\mathbb{E}_{\boldsymbol{v}}\left[L_{\mathscr{D}_{1} \cup \mathscr{D}_{2}}(\boldsymbol{w}+\Delta \boldsymbol{w} +\boldsymbol{v}) \right] = \mathbb{E}_{\boldsymbol{v}}\left[L_{\mathscr{D}_{1}}(\boldsymbol{w}+\Delta \boldsymbol{w}+\boldsymbol{v})\right]+\mathbb{E}_{\boldsymbol{v}}\left[L_{\mathscr{D}_{2}}(\boldsymbol{w}+\Delta \boldsymbol{w} +\boldsymbol{v})\right] \\
\leq \ &\underbrace{\mathbb{E}_{\boldsymbol{v}}\left[L_{\mathscr{D}_{1}}(\boldsymbol{w}+\Delta \boldsymbol{w} +\boldsymbol{v})\right]}_{\text {stability of old task}} +\underbrace{L_{\mathcal{D}_{2}}(\boldsymbol{w}+\Delta \boldsymbol{w})}_{\text {sensitivity of new task}}
+ \underbrace{\mathbb{E}_{\boldsymbol{v}}\left[L_{\mathcal{D}_{2}}(\boldsymbol{w}+\Delta \boldsymbol{w}+\boldsymbol{v}) \right] -L_{\mathcal{D}_{2}}(\boldsymbol{w}+\Delta \boldsymbol{w})}_{\text {expected sharpness on training set of new task}} \\
& + 4\sqrt{\frac{1}{n}\left(KL(\boldsymbol{w}+\Delta \boldsymbol{w}+\boldsymbol{v}||P)+\ln\frac{2n}{\delta}\right)},
\end{aligned}
\nonumber
\end{equation}
% \begin{equation}
% \begin{aligned}
% \min_{\Delta \boldsymbol{w}} &\mathbb{E}_{\boldsymbol{v}}\left[L_{\mathscr{D}_{1} \cup \mathscr{D}_{2}}(\boldsymbol{w}+\Delta \boldsymbol{w} +\boldsymbol{v}) \right] = \min_{\Delta \boldsymbol{w}}  \mathbb{E}_{\boldsymbol{v}}\left[L_{\mathscr{D}_{1}}(\boldsymbol{w}+\Delta \boldsymbol{w}+\boldsymbol{v})\right]+\mathbb{E}_{\boldsymbol{v}}\left[L_{\mathscr{D}_{2}}(\boldsymbol{w}+\Delta \boldsymbol{w} +\boldsymbol{v})\right] \\
% \leq \min_{\Delta \boldsymbol{w}} &\underbrace{\mathbb{E}_{\boldsymbol{v}}\left[L_{\mathscr{D}_{1}}(\boldsymbol{w}+\Delta \boldsymbol{w} +\boldsymbol{v})\right]}_{\text {stability of old task}} +\underbrace{L_{\mathcal{D}_{2}}(\boldsymbol{w}+\Delta \boldsymbol{w})}_{\text {sensitivity of new task}}
% + \underbrace{\mathbb{E}_{\boldsymbol{v}}\left[L_{\mathcal{D}_{2}}(\boldsymbol{w}+\Delta \boldsymbol{w}+\boldsymbol{v}) \right] -L_{\mathcal{D}_{2}}(\boldsymbol{w}+\Delta \boldsymbol{w})}_{\text {expected sharpness on training set of new task}} \\
% + &4\sqrt{\frac{1}{n}\left(KL(\boldsymbol{w}+\Delta \boldsymbol{w}+\boldsymbol{v}||P)+\ln\frac{2n}{\delta}\right)},
% \end{aligned}
% \end{equation}
where $n$ is the size of training set $\mathcal{D}_2$. As we can see, the PAC-Bayes bound in the continual learning scenario depends on four quantities, (1) the stability of old task, (2) the sensitivity of new task, (3) the expected sharpness on training set of new task, and (4) the Kullback Leibler (KL) divergence to the "prior" $P$. The bound is valid for any distribution measure
$P$, any perturbation distribution $\boldsymbol{v}$ and any method of choosing $\Delta \boldsymbol{w}$ dependent on the training set $\mathcal{D}_{2}$. 

In order to ensure the stability of old task, we constrain $\Delta \boldsymbol{w}$ in the complementary space $\mathcal{V}^C$ of the important space representing the old task $\mathscr{D}_1$ following GPM \cite{saha2021gradient}, so that  $\mathbb{E}_{\boldsymbol{v}}\left[L_{\mathscr{D}_{1}}(\boldsymbol{w} +\Delta \boldsymbol{w} +\boldsymbol{v})\right]$ does not increase too much compared with the previously minimized $\mathbb{E}_{\boldsymbol{v}}\left[L_{\mathscr{D}_{1}}(\boldsymbol{w}+\boldsymbol{v})\right]$. Let $\boldsymbol{v}\in\mathcal{V}$, we have $\mathbb{E}_{\boldsymbol{v} \in \mathcal{V}}\left[L_{\mathcal{D}_{2}}(\boldsymbol{w}+\Delta \boldsymbol{w}+\boldsymbol{v}) \right] \leq \max_{\boldsymbol{v} \in \mathcal{V}} L_{\mathcal{D}_{2}}(\boldsymbol{w}+\Delta \boldsymbol{w}+\boldsymbol{v})$, then we can rewrite the above bound as follows:

\begin{equation}
\textcolor{drd}{
\begin{aligned}
&\min_{\Delta \boldsymbol{w}} \mathbb{E}_{\boldsymbol{v}}\left[L_{\mathscr{D}_{1} \cup \mathscr{D}_{2}}(\boldsymbol{w}+\Delta \boldsymbol{w} +\boldsymbol{v}) \right] \leq \min_{\Delta \boldsymbol{w}\in  \mathcal{V}^C} \mathbb{E}_{\boldsymbol{v}}\left[L_{\mathscr{D}_{1}}(\boldsymbol{w} +\Delta \boldsymbol{w} +\boldsymbol{v})\right] + L_{\mathcal{D}_{2}}(\boldsymbol{w}+\Delta \boldsymbol{w}) \\
&+\max_{\boldsymbol{v}\in \mathcal{V}} L_{\mathcal{D}_{2}}(\boldsymbol{w}+\Delta \boldsymbol{w}+\boldsymbol{v}) -L_{\mathcal{D}_{2}}(\boldsymbol{w}+\Delta \boldsymbol{w}) +4\sqrt{\frac{1}{n}\left(KL(\boldsymbol{w}+\Delta \boldsymbol{w} +\boldsymbol{v}||P)+\ln\frac{2n}{\delta}\right)}.
\end{aligned}
}\nonumber
\end{equation}

% \begin{equation}
% \begin{aligned}
% \min_{\Delta \boldsymbol{w}} \mathbb{E}_{\boldsymbol{v} \in \mathcal{V}}\left[L_{\mathscr{D}_{1} \cup \mathscr{D}_{2}}(\boldsymbol{w}+\Delta \boldsymbol{w} +\boldsymbol{v}) \right] \leq \min_{\Delta \boldsymbol{w}\in  \mathcal{V}^C} \mathbb{E}_{\boldsymbol{v} \in \mathcal{V}}\left[L_{\mathscr{D}_{1}}(\boldsymbol{w} +\Delta \boldsymbol{w} +\boldsymbol{v})\right] + L_{\mathcal{D}_{2}}(\boldsymbol{w}+\Delta \boldsymbol{w}) \\
% +\max_{\boldsymbol{v}\in \mathcal{V}} L_{\mathcal{D}_{2}}(\boldsymbol{w}+\Delta \boldsymbol{w}+\boldsymbol{v}) -L_{\mathcal{D}_{2}}(\boldsymbol{w}+\Delta \boldsymbol{w}) +4\sqrt{\frac{1}{n}\left(KL(\boldsymbol{w}+\Delta \boldsymbol{w} +\boldsymbol{v}||P)+\ln\frac{2n}{\delta}\right)}.
% \end{aligned}
% \nonumber
% \end{equation}

Thus, our FS-DGPM exactly optimizes the worst-case of the flatness of the weight loss landscape to control the PAC-Bayes bound, which theoretically justifies both lower loss value and flatter landscape lead to better continual learning performance, and why our proposed FS-DGPM works.

\subsection{Derivation for DGPM}
\label{app_dev_weight}

We derive the gradient of the importance value $\lambda_i$ by minimizing the worst performance of $f_{\boldsymbol{w}+\boldsymbol{v}}$ under the current task batch $\hat{\mathcal{D}}_t$ and the batch $\hat{\mathcal{M}}$ sampled from the replay buffer $\mathcal{M}$:
\begin{equation}
\begin{aligned}
\nabla_{\lambda_i} L_{\hat{\mathcal{D}}_t \cup \hat{\mathcal{M}}}(\boldsymbol{w}+\boldsymbol{v}) =&\frac{\partial}{\partial \left(\boldsymbol{w}+\boldsymbol{v}\right) } L_{\hat{\mathcal{D}}_t \cup \hat{\mathcal{M}}}\left(\boldsymbol{w}+\boldsymbol{v}\right) \cdot \frac{\partial}{\partial \lambda_{i}} \left( \boldsymbol{w}+\boldsymbol{v}\right)\\
= &\frac{\partial}{\partial \left(\boldsymbol{w}+\boldsymbol{v}\right) } L_{\hat{\mathcal{D}}_t \cup \hat{\mathcal{M}}}\left(\boldsymbol{w}+\boldsymbol{v}\right) \cdot \frac{\partial}{\partial \lambda_{i}} \left( \boldsymbol{w}+ \eta_1 \sum_{j=1}^{k} \lambda_{j}\boldsymbol{u}_{j}\boldsymbol{u}_{j}^{T} \left( \nabla_{\boldsymbol{w}} L_{\hat{\mathcal{D}}_t}(\boldsymbol{w})\right)\right) \\
= &\frac{\partial}{\partial \left(\boldsymbol{w}+\boldsymbol{v}\right) } L_{\hat{\mathcal{D}}_t \cup \hat{\mathcal{M}}}\left(\boldsymbol{w}+\boldsymbol{v}\right) \cdot \left(\eta_1 \boldsymbol{u}_{i}\boldsymbol{u}_{i}^{T} \left( \nabla_{\boldsymbol{w}} L_{\hat{\mathcal{D}}_t}(\boldsymbol{w})\right)\right) \\
\approx &\eta_{1} \left( \nabla_{\boldsymbol{w}} L_{\hat{\mathcal{D}}_t \cup \hat{\mathcal{M}}}\left(\boldsymbol{w}\right) \right)^T \boldsymbol{u}_{i} \boldsymbol{u}_{i}^{T} \left( \nabla_{\boldsymbol{w}} L_{\hat{\mathcal{D}}_t}\left(\boldsymbol{w}\right)\right),
\end{aligned}
\nonumber
\end{equation}
where $\cdot$ is the dot product operator. Note that we only consider one gradient update to $\boldsymbol{v}$ in the second equation for simplicity, but using multiple gradient updates is a straightforward extension. For the third equation, we get it by assuming that $\boldsymbol{w}$ is constant with respect to $\lambda_i$. The last approximation is obtained by the first-order Taylor expansion. Setting all first-order gradient terms as constants to ignore second-order derivatives, we get the approximation as:
$$
\nabla_{(\boldsymbol{w}+\boldsymbol{v})} L_{\hat{\mathcal{D}}_t \cup \hat{\mathcal{M}}}(\boldsymbol{w}+\boldsymbol{v}) = \nabla_{\boldsymbol{w}}  L_{\hat{\mathcal{D}}_t \cup \hat{\mathcal{M}}}\left(\boldsymbol{w}\right) + \left(\nabla_{\boldsymbol{w}}^2 L_{\hat{\mathcal{D}}_t \cup \hat{\mathcal{M}}}\left(\boldsymbol{w}\right)\right) \boldsymbol{v} +  O\left(\|\boldsymbol{v}\|^2\right) \approx \nabla_{\boldsymbol{w}} L_{\hat{\mathcal{D}_t} \cup \hat{\mathcal{M}}}\left(\boldsymbol{w}\right).
$$
The importance of each basis is constrained to be between $0$ and $1$, where $0$ indicates that the basis is not important to old tasks and can completely release for learning new tasks. The initial value of all importance is set to $1$, and we use the sigmoid function with a temperature factor of $10$ \textcolor{drd}{at the end of gradient update}: $\lambda_i \leftarrow 1/(1+exp(-10\lambda_i))$.

\subsection{Pseudo-code for updating GPM}
\label{app_base}
% Recall that $\boldsymbol{w}$ and $\boldsymbol{v}$ are all layer-wise updated.
\textcolor{drd}{GPM \cite{saha2021gradient} achieves excellent stability by ensuring that gradient updates only occur in directions orthogonal to the gradient subspaces deemed important for the past tasks. Similar to \cite{saha2021gradient}, we calculate the bases of these subspaces for each layer by analyzing network representations after learning each task with Singular Value Decomposition (SVD), and then use it to update $\boldsymbol{v}$ and $\boldsymbol{w}$ by layer.}

% We calculate the bases matrix for each layer following GPM \cite{saha2021gradient}, and then use it to update $\boldsymbol{v}$ and $\boldsymbol{w}$ by layer. 
As shown in Algorithm \ref{alg_base} for updating GPM, we firstly sample $n_s$ random examples from the replay buffer $\mathcal{M}$ to construct the representation matrix for each layer, $\boldsymbol{R}^l$. Next, we perform SVD on $\boldsymbol{R}^l = \boldsymbol{U}^l \Sigma^{l} (\boldsymbol{V}^l )^T$ and obtained its $k$-rank approximation $\boldsymbol{R}^{l}_{k}$ according to the following \textbf{criteria} for the given threshold, $\epsilon^{l}$:
\begin{equation}
\label{svd}
\|\boldsymbol{R}^{l}_{k}\|_{F}^{2} \geq \epsilon^{l} \|\boldsymbol{R}^{l}\|_{F}^{2},
\end{equation}
where $\|\cdot\|_{F}$ is the Frobenius norm of the matrix and and $\epsilon^{l}$ $(0 < \epsilon^{l} \leq 1)$ is the threshold hyperparameter for layer $l$.

\begin{algorithm}[H]
	\caption{UpdateGPM} 
	\label{alg_base}
	\begin{algorithmic}
		\State  \textbf{Input:} Network $f_{\boldsymbol{w}}$ with $L$-layer, Replay buffer $\mathcal{M}$, sample size $n_s$, threshold $\epsilon$ for each layer. 
		\State  \textbf{Result:} Bases matrix $\{\left(\boldsymbol{M}_{l}\right)_{l=1}^{L}\}$ {\color{gray!66}$\quad$ \Comment{till $L-1$ if multi-head setting.}}
		\If{$\mathcal{M}$ is not empty}
		\State $B_{n_{s}} \sim \mathcal{M}$ {\color{gray!66}$\quad$ \Comment{sample a mini-batch of size $n_s$ from $\mathcal{M}$.}}
		\State $\mathcal{R} \leftarrow$ forward$\left(B_{n_{s}}, f_{\boldsymbol{w}}\right)$, where $\mathcal{R}=\{(\boldsymbol{R}^{l})_{l=1}^{L}\}$   {\color{gray!66}  \Comment{\parbox[t]{.41\linewidth}{construct representation matrix for each layer by forward pass.}}}
		\For{$l=1,2,\cdots,L$}
		\State $\boldsymbol{U}^{l} \leftarrow \operatorname{SVD}(\boldsymbol{R}^{l})$ {\color{gray!66} \Comment{update new bases for each layer by performing SVD.}}
		\State $k \leftarrow$ criteria $(\boldsymbol{R}^{l}, \epsilon^{l})$ {\color{gray!66} \Comment{see Eq. (\ref{svd}).}}
		\State $\boldsymbol{M}_{l} \leftarrow \boldsymbol{U}^{l}[0: k]$
		\EndFor
		\EndIf
	\end{algorithmic}
\end{algorithm}

For fully connected layer, the representation matrix $\boldsymbol{R}^l=[\boldsymbol{x}_{1}^{l}, \boldsymbol{x}_{2}^{l},\cdots,\boldsymbol{x}_{n_s}^{l}]$ concatenates $n_s$ inputs of the $l$-th layer linear function along the column, which are obtained by forwarding the batch of $n_s$ samples $\{\boldsymbol{x}_1,\cdots, \boldsymbol{x}_{n_s}\}$ through the network $f_{\boldsymbol{w}}$.
For convolution (Conv) layers, we first express the Conv as matrix multiplication by reshaping the input tensor $\mathcal{X} \in \mathbb{R}^{C_i \times h_i \times w_i}$ and filters $\mathcal{W} \in \mathbb{R}^{C_o \times C_i \times k \times k}$ into $\boldsymbol{X} \in \mathbb{R}^{(h_o \times w_o) \times (C_i \times k \times k)}$ and $\boldsymbol{W} \in \mathbb{R}^{(C_i \times k \times k) \times C_o}$ respectively, where $C_i (C_o)$ denotes the number of input (output) channels of the Conv layer, $h_i, w_i (h_o, w_o)$ represents the height and width of the input (output) feature maps and $k$ is the kernel size of the filters. Then we construct the representation matrix as $\boldsymbol{R}^l=[(\boldsymbol{X}_{1}^{l})^T, (\boldsymbol{X}_{2}^{l})^T,\cdots,(\boldsymbol{X}_{n_s}^{l})^T] \in \mathbb{R}^{(C_i \times k \times k) \times (h_o \times w_o \times n_s)}$. \textcolor{drd}{The key difference with GPM \cite{saha2021gradient} is that we perform SVD in the entire gradient space and use the obtained bases to replace the last calculated bases, while \cite{saha2021gradient} obtains the newly added bases by performing SVD in the subspace orthogonal to the existing bases. In addition, GPM can be regarded as a special case of our method when $\eta_1$ and $\eta_2$ are set to $0$.}

\section{Details for Landscape Visualization for Continual Learning}
\label{app_visual}
In this section, we first provide the pseudo-code of the visualization of the weight loss landscape in continual learning, and then provide more empirical results.

\subsection{Pseudo-code for Visualization}
\label{app_visual_setting}
As shown in Algorithm \ref{alg_visual} for the visualization of the weight loss landscape for the continual learning scenario, we first sample a random direction $\boldsymbol{d}$ from a Gaussian distribution, and then apply the filter-wise normalization following \cite{li2018visualizing} to eliminate the scaling invariance of DNNs. Next, we independently calculate the training loss of a series of perturbed weights for each learned task. For a given task descriptor $\tau$ and perturbed weights $\boldsymbol{w}+\alpha \boldsymbol{d}$, we obtain the training loss of the perturbed model $f_{\boldsymbol{w}+\alpha \boldsymbol{d}}$ on all training samples of task $\tau$. Then, we plot the weight loss landscape for task $\tau$. If the descriptor for the current training task of the model is $t$, we will plot $t$ curves.
\begin{algorithm}[H]
	\caption{Visualization of the Weight Loss Landscape} 
	\label{alg_visual}
	\begin{algorithmic}
		\State  \textbf{Input:} Network $f_{\boldsymbol{w}}$ with $L$-layer ($F_l$ filters in the $l$-th layer), current task descriptor $t$, training dataset $\mathcal{D}_{\tau}=\{\boldsymbol{x}_{i,\tau},y_{i,\tau}\}_{i=1}^{n_{\tau}}$ for $\tau=1,...,t$ , the scalar parameter $\alpha \in [\alpha_{min},\alpha_{max}]$. 
		\State Sample a random direction $\boldsymbol{d} \sim \mathcal{N}(0, 1)$ 
		\For{$l=1,2,\cdots,L$}
		\For{$j=1,2,\cdots,F_l$}
		\State $\boldsymbol{d}_{l,j} \leftarrow \frac{\boldsymbol{d}_{l,j}}{\|\boldsymbol{d}_{l,j}\|_F}\|\boldsymbol{w}_{l,j}\|_F$ {\color{gray!66}  \Comment{\parbox[t]{.55\linewidth}{filter-wise normalization for the random direction.}}}
		\EndFor
		\EndFor
		
		\For{$\tau=1,2,\cdots,t$}
		\For{$\alpha=\alpha_{min},\cdots,\alpha_{max}$}
		\State $L_{\mathcal{D}_{\tau}}(\boldsymbol{w}+\alpha \boldsymbol{d}) = \frac{1}{n_{\tau}}\sum_{\left(\boldsymbol{x}, y\right)\in \mathcal{D}_{\tau}} \ell \left(f_{\boldsymbol{w}+\alpha\boldsymbol{d}}\left(\boldsymbol{x} \right), y\right)$ {\color{gray!66}  \Comment{\parbox[t]{.3\linewidth}{calculate training loss of the perturbed model on task $\tau$.}}}
		\EndFor
		\State Plot$(\alpha, L_{\mathcal{D}_{\tau}}(\boldsymbol{w}+\alpha \boldsymbol{d}))$, $\forall \alpha \in [\alpha_{min},\alpha_{max}]$
		\EndFor
	\end{algorithmic}
\end{algorithm}

\subsection{More Results for Section \ref{subsec_visual}: Connection of Weight Loss Landscape and Continual Learning}
\label{app_visual_results}

\begin{figure}[t]
%\vspace{-4mm}
    \centering
    \includegraphics[width=1.0\textwidth]{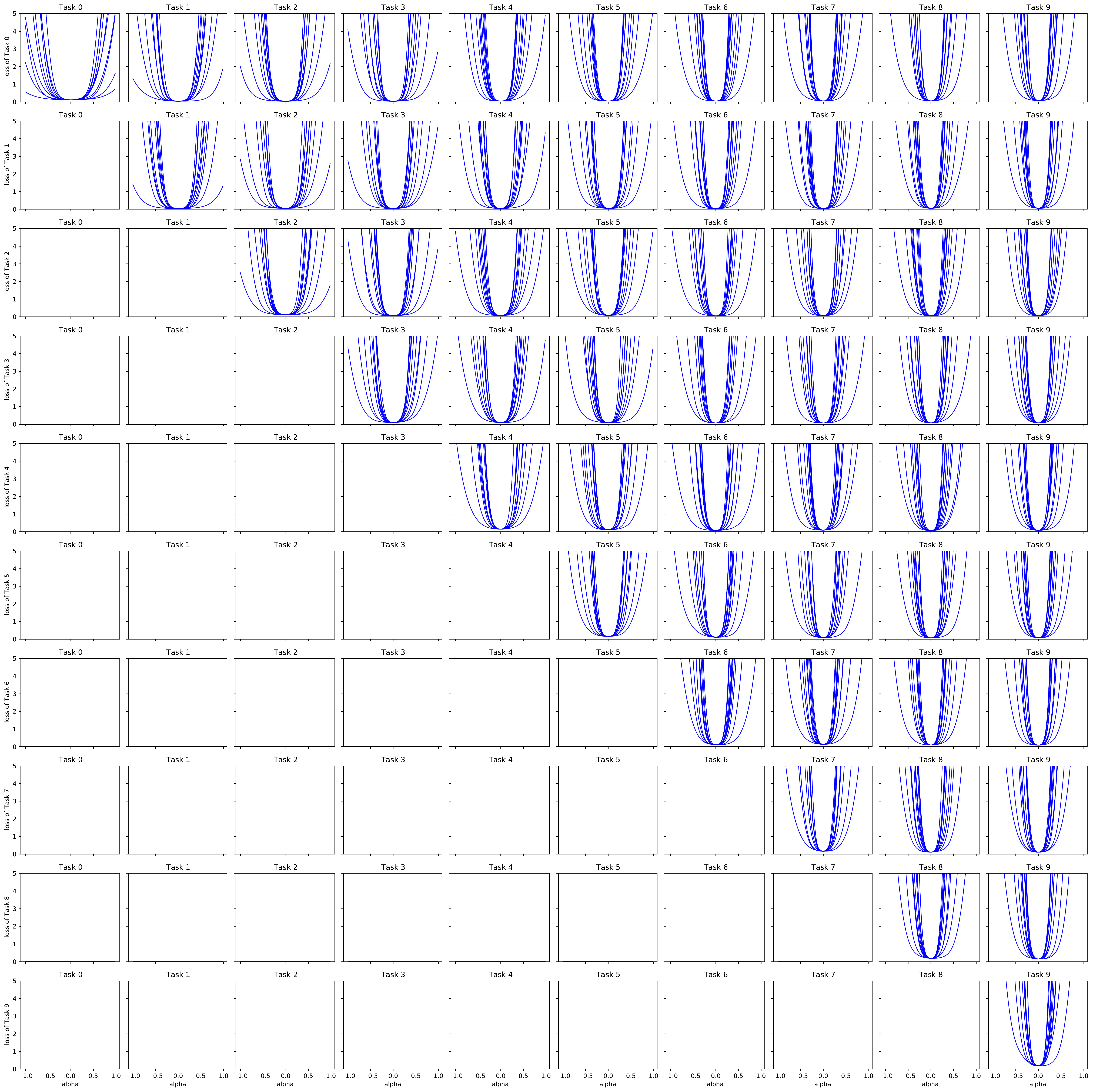}
    \caption{The weight loss landscape of the ten tasks of PMNIST using Oracle for training. The $i$-th row represents changes in the weight loss landscape of the $i$-th task during the model evolution, and the $j$-th column indicates that the model has learned $j$ tasks. The y-axis is the loss value, and the x-axis is the scalar value for visualization random direction. (Task $t$ is the abbr. of $t+1$-th task.)}
    \label{fig_app_multi}
\end{figure}

\begin{figure}[t]
%\vspace{-4mm}
    \centering
    \includegraphics[width=1.0\textwidth]{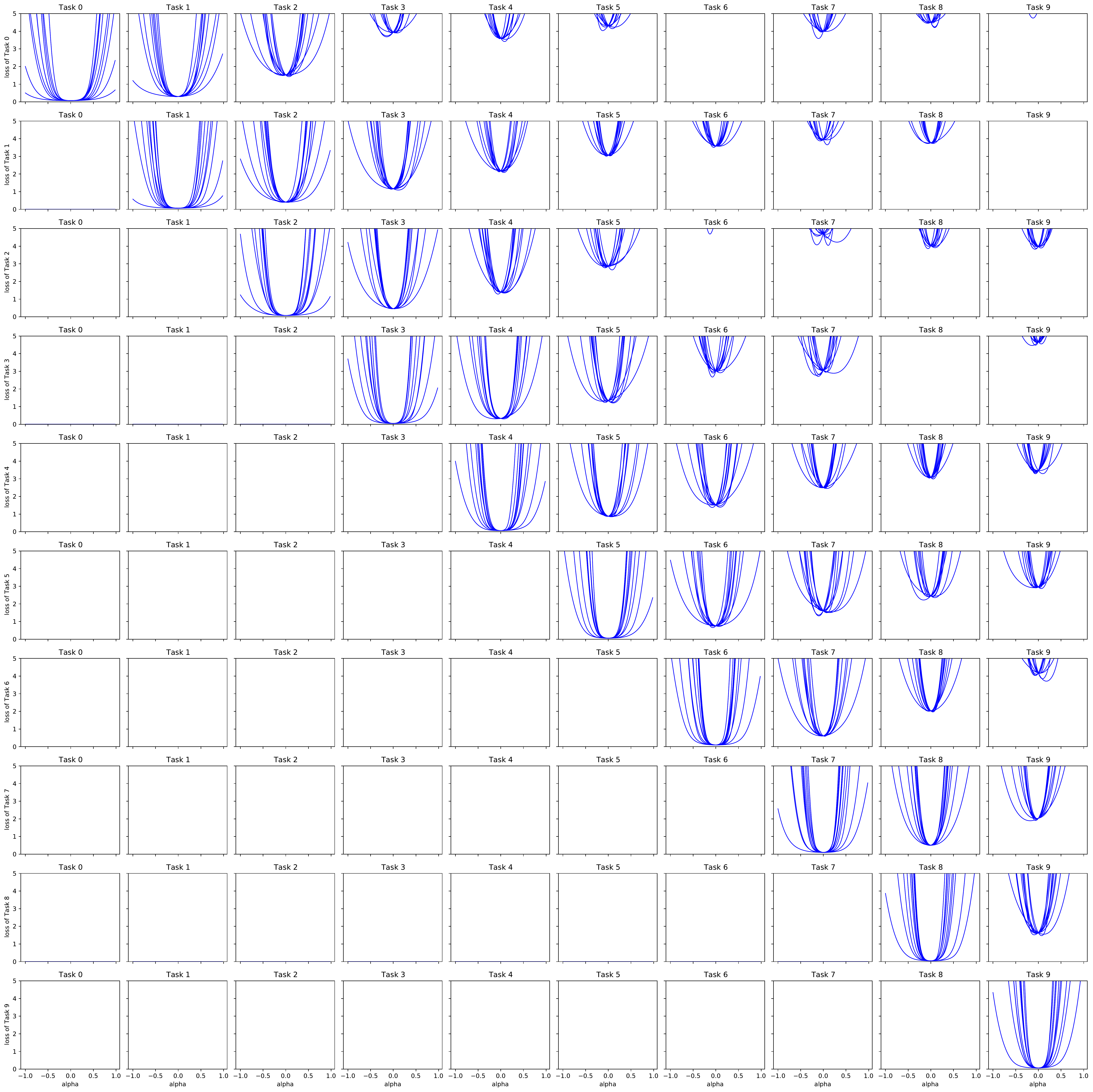}
    \caption{The weight loss landscape of the ten tasks of PMNIST using Finetune for training. The $i$-th row represents changes in the weight loss landscape of the $i$-th task during the model evolution, and the $j$-th column indicates that the model has learned $j$ tasks. The y-axis is the loss value, and the x-axis is the scalar value for visualization random direction. (Task $t$ is the abbr. of $t+1$-th task.)}
    \label{fig_app_finetune}
\end{figure}

\begin{figure}[t]
%\vspace{-4mm}
    \centering
    \includegraphics[width=1.0\textwidth]{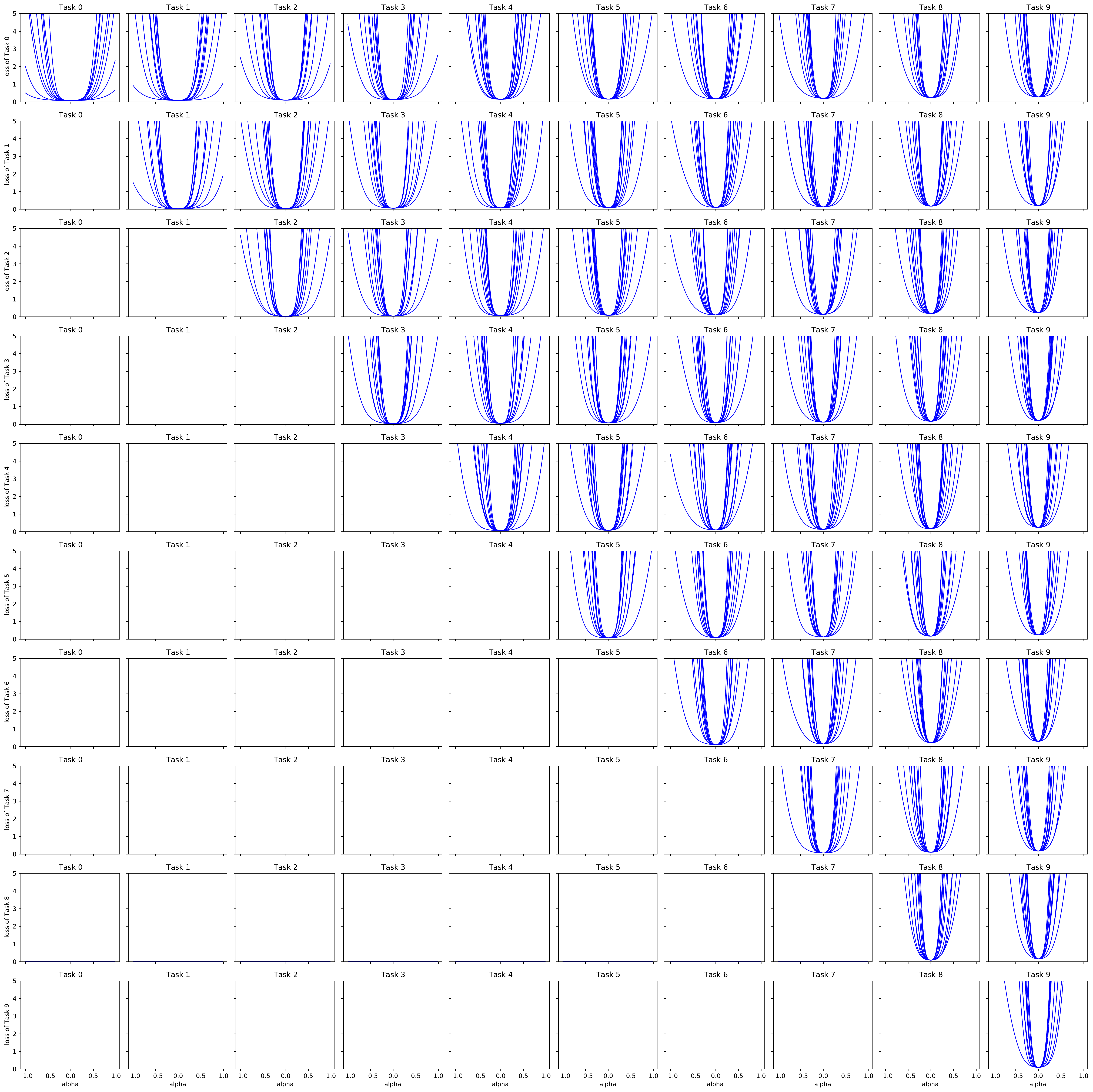}
    \caption{The weight loss landscape of the ten tasks of PMNIST using GPM for training. The $i$-th row represents changes in the weight loss landscape of the $i$-th task during the model evolution, and the $j$-th column indicates that the model has learned $j$ tasks. The y-axis is the loss value, and the x-axis is the scalar value for visualization random direction. (Task $t$ is the abbr. of $t+1$-th task.)}
    \label{fig_app_gpm}
\end{figure}

\begin{figure}[t]
%\vspace{-4mm}
    \centering
    \includegraphics[width=1.0\textwidth]{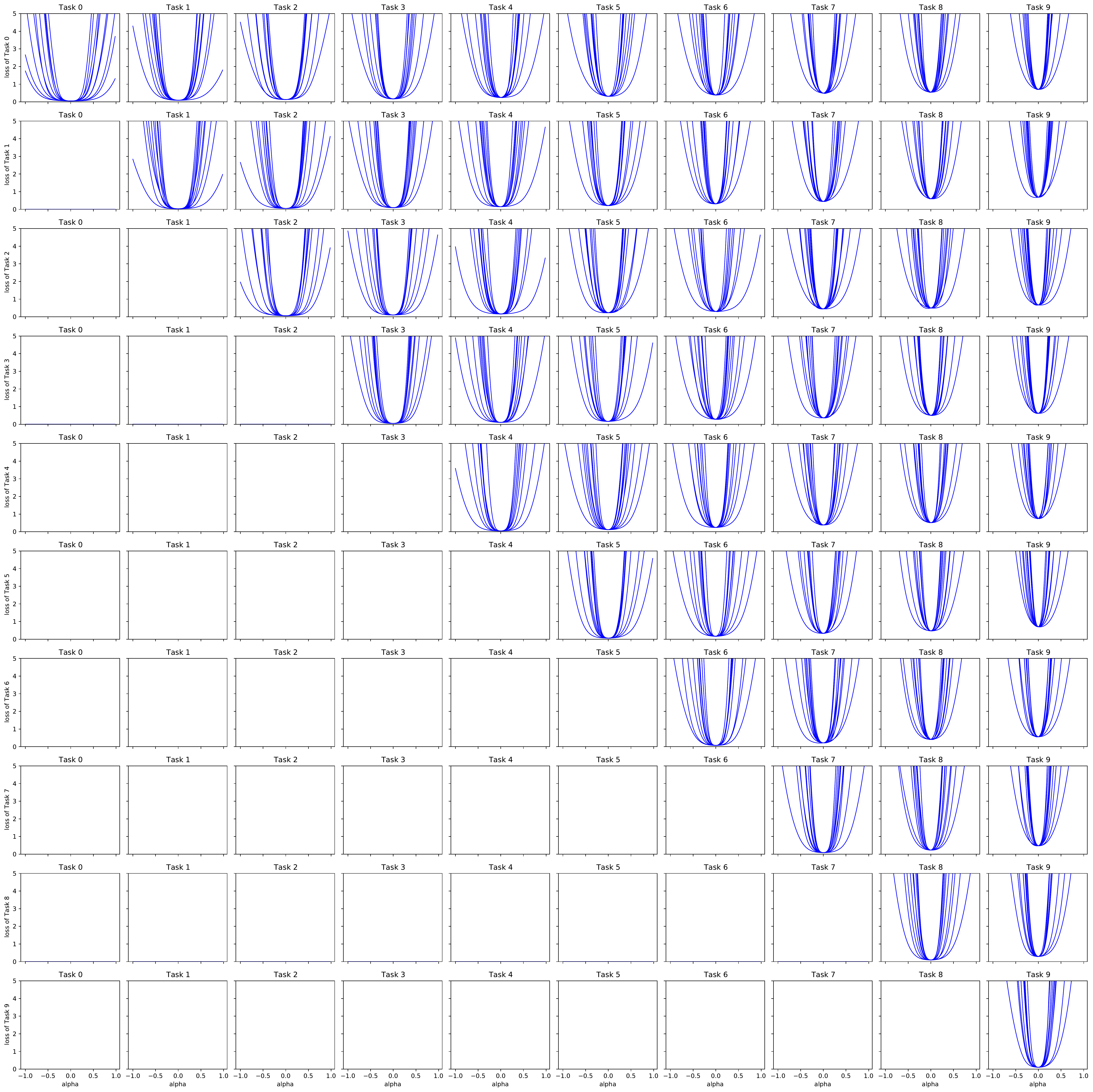}
    \caption{The weight loss landscape of the ten tasks of PMNIST using La-MAML for training. The $i$-th row represents changes in the weight loss landscape of the $i$-th task during the model evolution, and the $j$-th column indicates that the model has learned $j$ tasks. The y-axis is the loss value, and the x-axis is the scalar value for visualization random direction. (Task $t$ is the abbr. of $t+1$-th task.)}
    \label{fig_app_lamaml}
\end{figure}

To investigate the relationship between the weight loss landscape and stability-sensitivity in the continual learning scenario, we use the previously proposed GPM \cite{saha2021gradient}, La-MAML \cite{gupta2020maml}, and ER \cite{chaudhry2019continual} to train an MLP network with two hidden layers on the Permuted MNIST (PMNIST) \cite{lecun1998gradient} dataset that contains $10$ tasks. \textcolor{drd}{For each task, we use $60,000$ training samples instead of $1,000$ used in the experimental environment. The replay buffer size is set to $1,000$. We also use Oracle and Finetune, which respectively represent retraining the network on the entire dataset contain all passed tasks, and training the network on the data stream without any regularization or episodic memory.} Considering the direction $\boldsymbol{d}$ for visualization is randomly selected, we repeat the visualization 10 times with different $\boldsymbol{d}$. \textcolor{drd}{Figure \ref{fig_app_multi}, \ref{fig_app_finetune}, \ref{fig_app_gpm}, \ref{fig_app_lamaml}, and \ref{fig_app_er} show the weight loss landscape for each task when a new task is trained using Oracle, Finetune, GPM, La-MAML, and ER, respectively}. The $i$-th row of each figure represents changes in the weight loss landscape of the $i$-th task during the model evolution, and each $j$-th column indicates that the current model has learned $j$ tasks.

\section{Details for FS-ER}
\label{app_fser}
In this section, we first provide the pseudo-code of Flattening Sharpness for Vanilla ER, and then provide more empirical results.

\subsection{Pseudo-code for FS-ER}
\label{app_fser_alg}

Comparing the pseudo-code of Vanilla ER, FS-ER only adds the adversarial weight perturbation $\boldsymbol{v}$.
\begin{algorithm}[H]
	\caption{FS-ER} 
	\label{alg_fser}
	\begin{algorithmic}
		\State  \textbf{Input:} Network weight $\boldsymbol{w}$, loss function $\ell$, learning rate $\eta_3$, FS step size $\eta_1$, FS steps $K$, batch size $b$.
		\State  Initializing $\mathcal{M} \leftarrow\{\} $
		\For{$t=1,2,\cdots,T$}
		\For{$ep=1,2,\cdots,num_{epochs}$}
		\For{batch $\hat{\mathcal{D}}_t \stackrel{b}{\sim} \mathcal{D}_t$} {\color{gray!66}  \Comment{\parbox[t]{.55\linewidth}{Sample without replacement a mini-batch from task $t$.}}}
		\State $\hat{\mathcal{M}} \stackrel{b}{\sim} \mathcal{M}$ {\color{gray!66}  \Comment{\parbox[t]{.32\linewidth}{Sample a mini-batch from $\mathcal{M}$.}}}
		\For{$k=1,\cdots,K$}
		\State $\boldsymbol{v} \leftarrow \boldsymbol{v} + \eta_{1} \nabla_{(\boldsymbol{w}+\boldsymbol{v})} L_{\hat{\mathcal{D}}_t \cup \hat{\mathcal{M}}}(\boldsymbol{w}+\boldsymbol{v})$ {\color{gray!66}  \Comment{\parbox[t]{.4\linewidth}{Add adversarial weight perturbation $\boldsymbol{v}$.}}}
		\EndFor
		\State $\boldsymbol{w} \leftarrow \boldsymbol{w} -\eta_3 \nabla_{\boldsymbol{w}} L_{\hat{\mathcal{D}}_t \cup \hat{\mathcal{M}}}(\boldsymbol{w}+\boldsymbol{v})$ 
		\State Push $\hat{\mathcal{D}}_t$ to $\mathcal{M}$ with reservior sampling
		\EndFor
		\EndFor
		\EndFor
	\end{algorithmic}
\end{algorithm}

\subsection{More Results for Section \ref{subsec_fser}: A Case Study of Flattening Sharpness for Vanilla ER}
\label{app_fser_results}

\begin{figure}[t]
%\vspace{-4mm}
    \centering
    \includegraphics[width=1.0\textwidth]{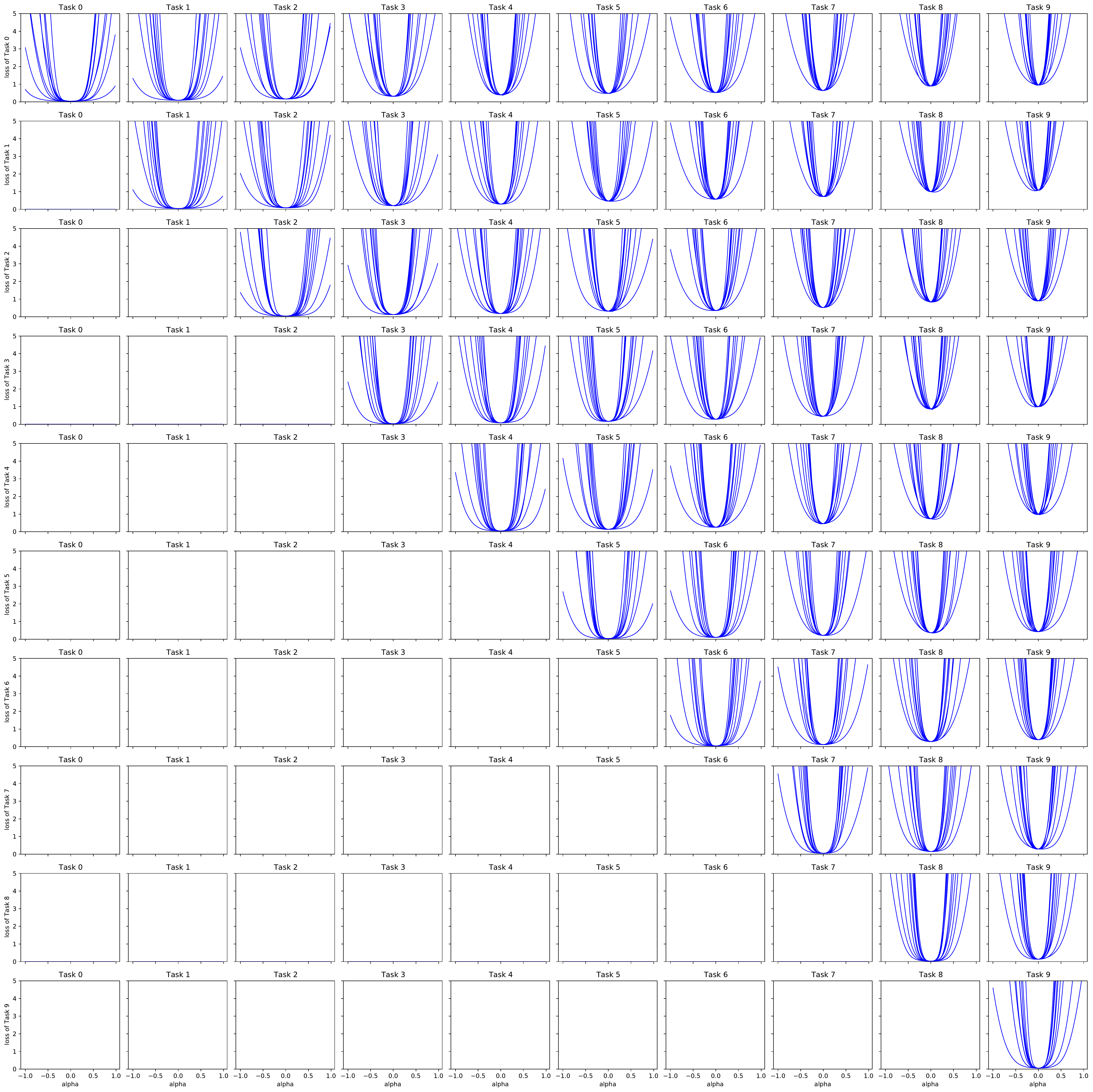}
    \caption{The weight loss landscape of the ten tasks of PMNIST using ER for training. The $i$-th row represents changes in the weight loss landscape of the $i$-th task during the model evolution, and the $j$-th column indicates that the model has learned $j$ tasks. The y-axis is the loss value, and the x-axis is the scalar value for visualization random direction. (Task $t$ is the abbr. of $t+1$-th task.)}
    \label{fig_app_er}
\end{figure}

\begin{figure}[t]
%\vspace{-4mm}
    \centering
    \includegraphics[width=1.0\textwidth]{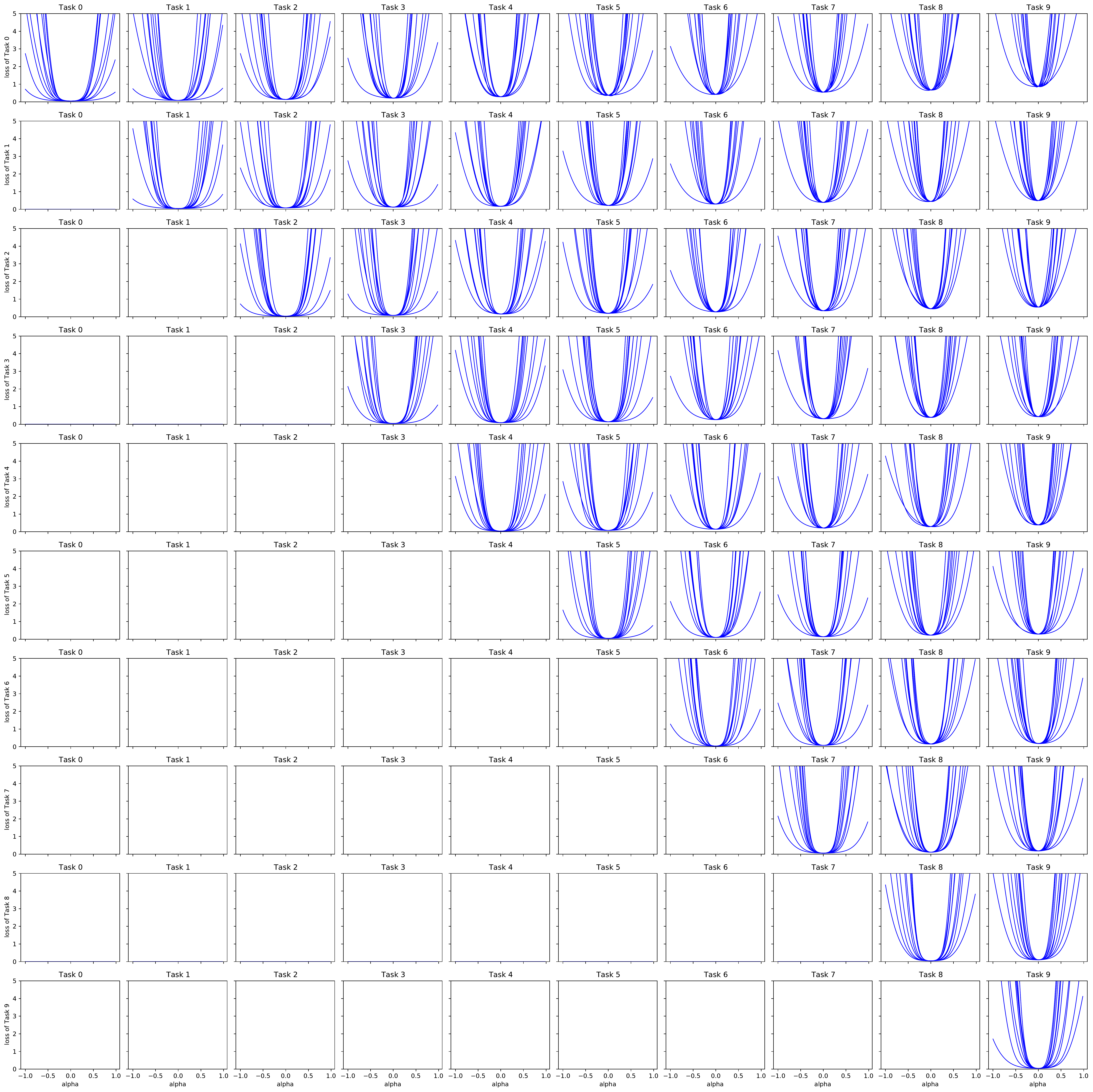}
    \caption{The weight loss landscape of the ten tasks of PMNIST using FS-ER for training. The $i$-th row represents changes in the weight loss landscape of the $i$-th task during the model evolution, and the $j$-th column indicates that the model has learned $j$ tasks. The y-axis is the loss value, and the x-axis is the scalar value for visualization random direction. (Task $t$ is the abbr. of $t+1$-th task.)}
    \label{fig_app_fser}
\end{figure}

Figure \ref{fig_app_er} and \ref{fig_app_fser} show the weight loss landscape for each task when a new task is trained using ER and FS-ER, respectively.

\section{Experimental Details}
\label{app_exp}

\subsection{Datasets}
\label{app_dataset}

Table \ref{app_table_dataset} summarizes the statistics of four datasets used in our experiments.

\begin{table}[ht]
	\caption{Dataset Statistics.}
	\label{app_table_dataset}
	\vspace{-4mm}
	\begin{center}
	\resizebox{\textwidth}{!}{
		\begin{tabular}{lcccc}
			\toprule
			&\textbf{PMNIST} &\textbf{CIFAR-100 Split}  &\textbf{CIFAR-100 Superclass} &\textbf{TinyImageNet} \\ \midrule
			Input size   & 1$\times$28$\times$28 & 3$\times$32$\times$32  & 3$\times$32$\times$32 & 3$\times$64$\times$64 \\
			\# tasks                 & 20 & 10 & 20 & 40 \\
			\# Classes/task             	 & 10 & 10 & 5  & 5 \\
			\# Training/task        & 1,000 & 4,750 & 2,375  & 2,250 \\
			\# Validation/task         		&- & 250 & 125 & 250 \\ 
			\# Test/task  & 10,000 & 1,000 & 500 & 250 \\
			\bottomrule
		\end{tabular}}
	\end{center}
\end{table}

\subsection{Architecture}
\label{app_arch}

\paragraph{AlexNet-like architecture:}
For 10-Split CIFAR-100, similar to GPM \cite{saha2021gradient}, we use a AlexNet-like architecture with three convolutional (Conv) layers and two fully connected layers. The three Conv layers have 64, 128, and 256 filters with 4$\times$4, 3$\times$3, and 2$\times$2 kernel sizes, respectively. Both fully connected layers have 2048 units in each layer. Max-pooling layer with filters of size 2$\times$2 is used after the Conv layer. Dropout of 0.2 is used for the first two layers, and 0.5 is used for the remaining layers. Batch normalization is only used in the second layer. 

\paragraph{Modified Lenet-5 architecture:}
For 20-Spilt CIFAR-100 Superclass, similar to GPM \cite{saha2021gradient}, we use a modified LeNet-5 architecture with 20-50-800-500 neurons, of which the first two are Conv layers with 5$\times$5 kernel sizes, and the last two are fully connected layers. Batch normalization and max-pooling layer with filters of size 3$\times$3 with a stride of $2$ are used in the Conv layers. \textcolor{drd}{Batch normalization parameters are learned for the first task and shared with all the other tasks.}

\paragraph{Architecture for TinyImageNet:}
For 40-Spilt TinyImageNet, similar to La-MAML \cite{gupta2020maml}, we use a CNN having 4 Conv layers with 160 3$\times$3 filters. The output from the final Conv layer is flattened and is passed through 2 fully connected layers having 320 and 640 units, respectively. 

All networks use ReLU in the hidden units, and finally have a multi-head output layer to perform classification for every task. No bias units are used following \cite{saha2021gradient}.

\subsection{Baselines}
\label{app_baseline}
We compare our method against multiple methods described below. 
\begin{itemize}
\item \textbf{EWC} \cite{kirkpatrick2017overcoming}: Elastic Weight Consolidation is a regularization-based method that uses the diagonal of Fisher information to identify important weights.
\item \textbf{ICARL} \cite{rebuffi2017icarl}: ICARL is a memory-based method that uses knowledge distillation and episodic memory to reduce forgetting.
\item \textbf{GEM} \cite{lopez2017gradient}: Gradient Episodic Memory uses the gradient of episodic memory to constrain optimization to make sure that the gradients of the new task do not change the previous knowledge.
\item \textbf{ER} \cite{riemer2018learning, chaudhry2019continual}: Experience Replay uses a small replay buffer to store old data using reservoir sampling.  Then, the stored data is replayed again with the new data samples. 
\item \textbf{La-MAML} \cite{gupta2020maml}: Look-ahead MAML is inspired by optimization-based meta-learning that leverages replay to avoid forgetting and favor positive backward transfer by asynchronously learning the weights and LRs. 
\item \textbf{GPM} \cite{saha2021gradient}: Gradient Projection Memory minimizes forgetting by taking gradient steps orthogonal to the gradient subspace deemed important for the past tasks when learning a new task.
\item \textbf{Multitask}: Multitask is an oracle baseline that all tasks are learned jointly using the entire dataset at once in a single network. Multitask is not a continual learning strategy but serves as an upper bound on average test accuracy on all tasks. 
\end{itemize}

GEM \cite{lopez2017gradient}, ICARL \cite{rebuffi2017icarl}, and La-MAML \cite{gupta2020maml} are implemented from the official implementation provided by \cite{gupta2020maml}. EWC \cite{kirkpatrick2017overcoming} is implemented from the official implementation provided by \cite{lopez2017gradient}. GPM \cite{saha2021gradient} is implemented from its official implementation. ER \cite{riemer2018learning, chaudhry2019continual} is implemented by adapting the code provided by \cite{riemer2018learning}.

\subsection{GPU Device}
\label{app_time}
We measured the average training time per task calculated on the NVIDIA GeForce RTX 2080 Ti GPU. Figure \ref{fig_exp}(b) shows the training time per task on the 20-Split CIFAR-100 Superclass experiment using the baselines and our method.

\subsection{Threshold Hyperparameter}
\label{app_thres}
Following \cite{saha2021gradient}, we use a different value of the threshold hyperparameter, $\epsilon$, for different architectures and different datasets. For PMNIST experiment, we use $\epsilon$ = 0.99 for all the layers and increasing the value of $\epsilon$ by 0.0005 for each new task. For CIFAR-100 Split and CIFAR-100 Superclass experiment, we use the values reported by \cite{saha2021gradient}. For CIFAR-100 Split, the initial value of $\epsilon$ is 0.97 for all the layers and increasing by 0.003 for each new task. For CIFAR-100 Superclass experiment, we use $\epsilon$ = 0.98 for all the layers and increasing by 0.001 for each new task. For TinyImageNet experiment, we use $\epsilon$ = 0.9 for all the layers and increasing by 0.0025 for each new task.

\subsection{List of Hyperparameters}
\label{app_hyperpara}

We report in Table \ref{app_table_hpyer} the hyper-parameters selected by grid-search for all baselines and our method. \textcolor{drd}{For PMNIST, we use a hyper-parameter called glances for all compared approaches and set it to $5$ following \cite{gupta2020maml}. This hyper-parameters indicates the number of meta-updates made on each incoming sample of data. In addition, for each task in PMNIST, we use 5 epochs to train the network for Multitask instead of 1 epoch in other methods.}

\begin{table}[ht]
	\caption{List of hyperparameters for the baselines and our approach. "LR" denotes the (initial) learning rate. Superclass is the abbr. of CIFAR-100 Superclass.}
	\label{app_table_hpyer}
	\vspace{-4mm}
	\begin{center}
	\resizebox{\textwidth}{!}{
		\begin{tabular}{lccccc}
			\toprule
			Method  &Parameter &\textbf{PMNIST} &\textbf{CIFAR-100 Split}  &\textbf{ Superclass} &\textbf{TinyImageNet} \\ \midrule
			EWC     & LR   & 0.01 & 0.005 & 0.03 & - \\
			        & memory strength, $\gamma$   & 100 & 5000 & 1000 & - \\
			\midrule
			ICARL 	& LR   & - & 0.03 & 0.01 & - \\
			        & memory strength, $\gamma$   & - & 0.1 & 0.5 & - \\
			        & memory size   & - & 1000 & 1000 & - \\
			\midrule
			GEM     & LR   & 0.01 & 0.01 & 0.03 & - \\
			        & memory strength, $\gamma$   & 0.0 & 0.5 & 0.5 & - \\
			        & memory size   & 200 & 1000 & 1000 & - \\
			\midrule
			ER  & LR   & 0.005 & 0.03 & 0.01 & - \\
			    & memory size   & 200 & 1000 & 1000 & - \\
			\midrule
			La-MAML	& LRs, $\alpha_0$   & 0.15 & 0.1 & 0.1 & - \\
			        & LR for LRs, $\eta$   & 0.3 & 0.5 & 0.5 & - \\
			        & memory size   & 200 & 1000 & 1000 & - \\
			\midrule
			GPM  & LR   & 0.01 & 0.01 & 0.01 & 0.005 \\
			     & $n_s$   & 300 & 125 & 125 & 200 \\
			\midrule
			Multitask & LR   & 0.01 & 0.01 & - & -  \\
			\midrule
			FS-DGPM & LR, $\eta_3$   & 0.01 & 0.01 & 0.01 & 0.01  \\
			        & LR for sharpness, $\eta_1$  & 0.05 & 0.001 & 0.01 & 0.001 \\
			        & LR for DGPM, $\eta_2$    & 0.01 & 0.01 & 0.01 & 0.01 \\
			        & memory size   & 200 & 1000 & 1000 & 400 \\
			        & $n_s$   & 200 & 125 & 125 & 200 \\
			\bottomrule
		\end{tabular}}
	\end{center}
\end{table}

\section{Additional Experimental Results}
\label{app_exp_result}

\begin{figure}[htbp]
\centering
\includegraphics[width=0.7\textwidth]{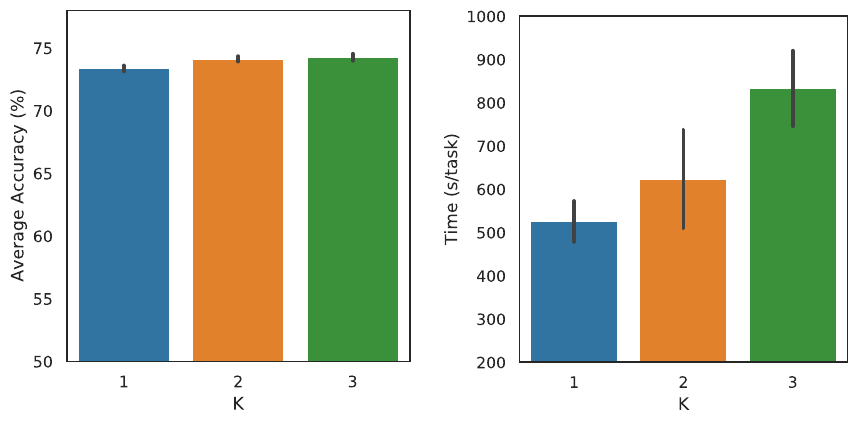}
\vspace{-1mm}
\caption{Results of different step numbers $K$ for weight perturbation $\boldsymbol{v}$ on CIFAR-100 Split in 50 epochs. Each experiment is run with 5 seeds.}
\label{fig_app_k}
\vspace{5mm}
\centering
\includegraphics[width=0.66\textwidth]{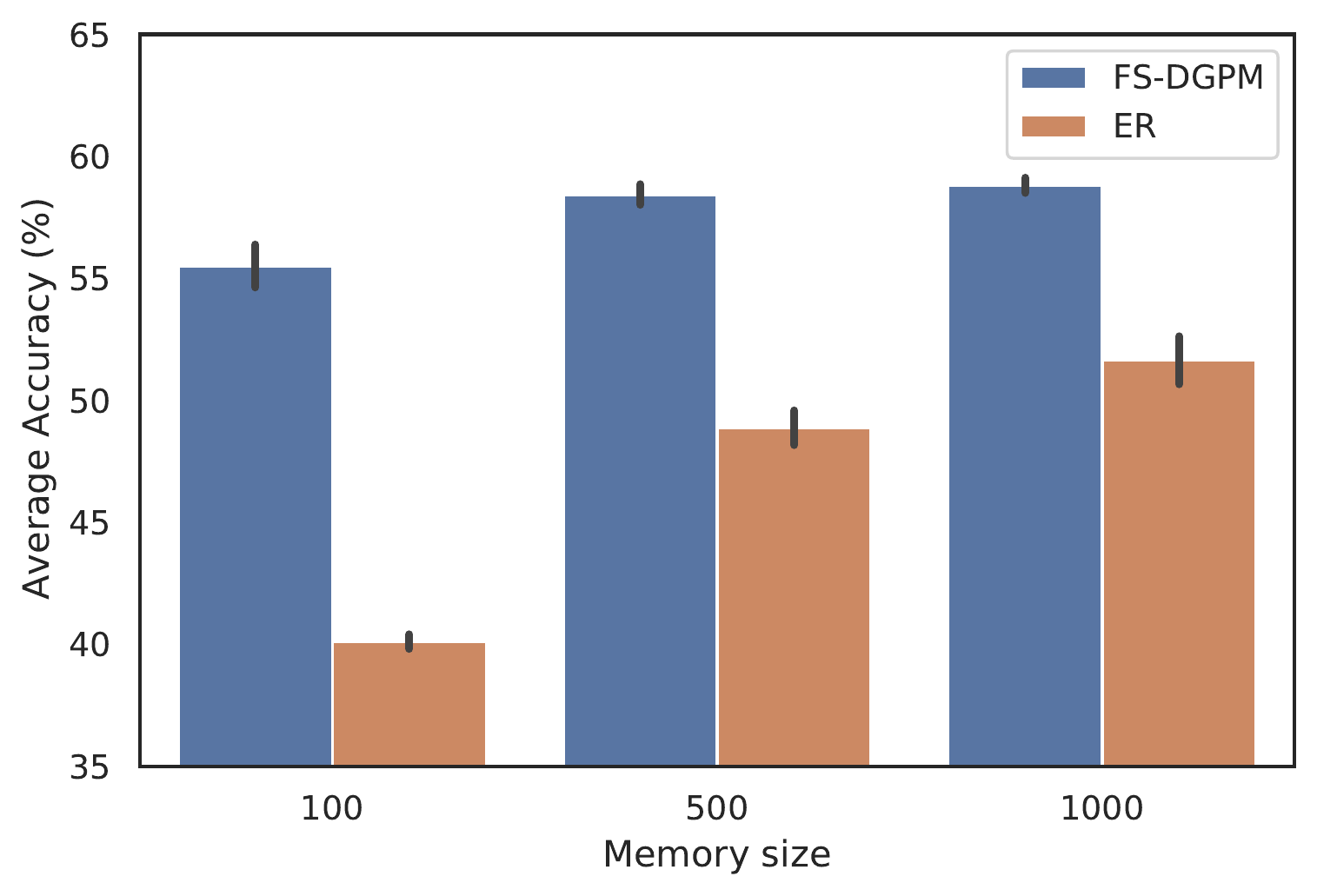}
\vspace{-1mm}
\caption{Average test accuracy of different memory size on CIFAR-100 SuperClass. Each experiment is run with 5 seeds. }
\label{fig_app_mem}
\vspace{5mm}
\centering
\includegraphics[width=0.66\textwidth]{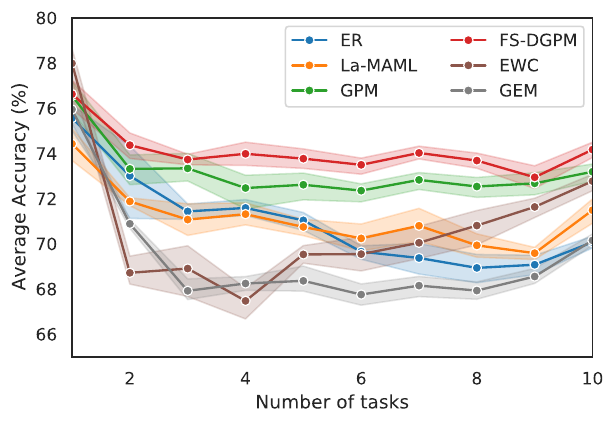}
\vspace{-1mm}
\caption{Average test accuracy as a function of the number of tasks trained on 10-Split CIFAR-100. Each experiment is run with 5 seeds. }
\label{fig_app_cifar}
\end{figure}

We provide the results of different step numbers $K$ in solving weight perturbation $\boldsymbol{v}$. We evaluation FS-DGPM with $K \in \{1,2,3\}$ in the CIFAR-100 Split experiment. As shown in Figure \ref{fig_app_k}, two steps have been well improved, and the extra steps only bring few improvements but with much more time. 

\textcolor{drd}{We also compare ER and our method on CIFAR-100 Superclass when the memory size is 100, 500, and 1000. As shown in Figure \ref{fig_app_mem}, we see that both ER and FS-DGPM benefits from increases in memory size, but the outperformance of FS-DGPM is more visible under the low-resource regime. We think the advantageous performance of FS-DGPM can be attributed to the effective utilization of episodic memory converted into bases through SVD.}

% \vspace{-1mm}

\clearpage
\section*{Checklist}

% %%% BEGIN INSTRUCTIONS %%%
% The checklist follows the references.  Please
% read the checklist guidelines carefully for information on how to answer these
% questions.  For each question, change the default \answerTODO{} to \answerYes{},
% \answerNo{}, or \answerNA{}.  You are strongly encouraged to include a {\bf
% justification to your answer}, either by referencing the appropriate section of
% your paper or providing a brief inline description.  For example:
% \begin{itemize}
%   \item Did you include the license to the code and datasets? \answerYes
%   \item Did you include the license to the code and datasets? \answerNo{The code and the data are proprietary.}
%   \item Did you include the license to the code and datasets? \answerNA{}
% \end{itemize}
% Please do not modify the questions and only use the provided macros for your
% answers.  Note that the Checklist section does not count towards the page
% limit.  In your paper, please delete this instructions block and only keep the
% Checklist section heading above along with the questions/answers below.
% %%% END INSTRUCTIONS %%%

\begin{enumerate}

\item For all authors...
\begin{enumerate}
  \item Do the main claims made in the abstract and introduction accurately reflect the paper's contributions and scope?
    \answerYes{}
  \item Did you describe the limitations of your work?
    \answerYes{Refer to Section \ref{sec_conclu} for details.}
  \item Did you discuss any potential negative societal impacts of your work?
    \answerNA{}
  \item Have you read the ethics review guidelines and ensured that your paper conforms to them?
    \answerYes{}
\end{enumerate}

\item If you are including theoretical results...
\begin{enumerate}
  \item Did you state the full set of assumptions of all theoretical results?
    \answerNA{}
	\item Did you include complete proofs of all theoretical results?
    \answerNA{}
\end{enumerate}

\item If you ran experiments...
\begin{enumerate}
  \item Did you include the code, data, and instructions needed to reproduce the main experimental results (either in the supplemental material or as a URL)?
    \answerYes{We provide the GitHub link of our code in the camera-ready version.}
  \item Did you specify all the training details (e.g., data splits, hyperparameters, how they were chosen)?
    \answerYes{See Appendix \ref{app_exp} for more details.}
	\item Did you report error bars (e.g., with respect to the random seed after running experiments multiple times)?
    \answerYes{}
	\item Did you include the total amount of compute and the type of resources used (e.g., type of GPUs, internal cluster, or cloud provider)?
    \answerYes{See Appendix \ref{app_time} for more details.}
\end{enumerate}

\item If you are using existing assets (e.g., code, data, models) or curating/releasing new assets...
\begin{enumerate}
  \item If your work uses existing assets, did you cite the creators?
    \answerYes{}
  \item Did you mention the license of the assets?
    \answerYes{}
  \item Did you include any new assets either in the supplemental material or as a URL?
    \answerYes{We provide the GitHub link of our code in the camera-ready version.}
  \item Did you discuss whether and how consent was obtained from people whose data you're using/curating?
    \answerNA{}
  \item Did you discuss whether the data you are using/curating contains personally identifiable information or offensive content?
    \answerNA{}
\end{enumerate}

\item If you used crowdsourcing or conducted research with human subjects...
\begin{enumerate}
  \item Did you include the full text of instructions given to participants and screenshots, if applicable?
    \answerNA{}
  \item Did you describe any potential participant risks, with links to Institutional Review Board (IRB) approvals, if applicable?
    \answerNA{}
  \item Did you include the estimated hourly wage paid to participants and the total amount spent on participant compensation?
    \answerNA{}
\end{enumerate}

\end{enumerate}

\end{document}